\documentclass[runningheads]{llncs}

\usepackage{eccv}

\usepackage{eccvabbrv}

\usepackage{graphicx}
\usepackage{booktabs}

\usepackage[accsupp]{axessibility}  % Improves PDF readability for those with disabilities.

\usepackage[pagebackref,breaklinks,colorlinks,citecolor=blue, linkcolor=red,urlcolor=blue]{hyperref}

\usepackage{orcidlink}

\usepackage{soul}  
\usepackage{gensymb} 
\usepackage{array} 
\usepackage{tikz, pgfplots}
\usepackage{multirow}
\usepackage{longtable}
\usepackage{colortbl}
\usepackage{wasysym}
\usepackage{wrapfig}

\usepackage{minitoc}

\newcommand{\ppcnt}{264~}
\newcommand{\hourcnt}{300~} 
\newcommand{\seqcnt}{1200~}  
   
\newcommand{\scriptcnt}{20~}
\newcommand{\loccnt}{50~}
\newcommand{\abcnt}{47~}
\newcommand{\multifloor}{31~}

\newcommand{\rooms}{201~}
\newcommand{\gardens}{45~}

\newcommand{\headtl}{399Km~} 
\newcommand{\wristtl}{1053Km~}
\newcommand{\posecnt}{260M~}

\newcommand{\imgcnt}{201.2M~}
\newcommand{\gazecnt}{10.8M~} 
\newcommand{\imucnt}{11.7B~} 

\newcommand{\motioncnt}{38.6 hours~} 
\newcommand{\atomiccnt}{207 hours~} 
\newcommand{\activitycnt}{196 hours~} 
\newcommand{\sentcnt}{310.5K~}
\newcommand{\wordcnt}{8.64M~}
\newcommand{\vocsz}{6545~}

\newcommand{\nymeria}{Nymeria~}

\begin{document}

\title{Nymeria: A Massive Collection of Multimodal Egocentric Daily Motion in the Wild} 

\titlerunning{Nymeria Dataset}

\author{Lingni Ma\inst{1} \and 
Yuting Ye\inst{1} \and
Fangzhou Hong\inst{2\dagger} \and 
Vladimir Guzov\inst{3\dagger} \and
Yifeng Jiang\inst{4\dagger} \and \\
Rowan Postyeni\inst{1} \and 
Luis Pesqueira\inst{1} \and 
Alexander Gamino\inst{1} \and 
Vijay Baiyya\inst{1} \and \\
Hyo Jin Kim\inst{1} \and 
Kevin Bailey\inst{1} \and 
David Soriano Fosas\inst{1} \and 
C. Karen Liu\inst{4} \and \\ 
Ziwei Liu\inst{2} \and 
Jakob Engel\inst{1} \and 
Renzo De Nardi\inst{1} \and 
Richard Newcombe\inst{1}
}

\authorrunning{Lingni Ma et al.}

\institute{Meta Reality Labs Research \and 
Nanyang Technological University, Singapore\and 
University of Tübingen and Max Planck Institute for Informatics, Germany\and
Stanford University, USA \\ 
\url{https://www.projectaria.com/datasets/nymeria}
}

\maketitle

\input{fig_teaser}
\begin{abstract}
We introduce \nymeria - a large-scale, diverse, richly annotated human motion dataset collected in the wild with multiple multimodal egocentric devices. The dataset comes with a) full-body ground-truth motion; b) multiple multimodal egocentric data from Project Aria devices with videos, eye tracking, IMUs and etc; and c) an third-person perspective by an additional ``observer''. All devices are precisely synchronized and localized in one metric 3D world. We derive hierarchical protocol to add in-context language descriptions of human motion, from fine-grain motion narrations, to simplified atomic actions and high-level activity summarization. To the best of our knowledge, \nymeria dataset is the world's largest human motion in the wild; first of its kind to provide synchronized and localized multi-device multimodal egocentric data; and the world’s largest motion-language dataset. It provides \hourcnt hours of daily activities from \ppcnt participants across \loccnt locations, total travelling distance over \headtl. The language descriptions contain \sentcnt sentences in \wordcnt words from a vocabulary size of 6545. To demonstrate the potential of the dataset we evaluate several SOTA algorithms for egocentric body tracking, motion synthesis, and action recognition.


\end{abstract}

\section{Introduction}
The advent of AI is leading to a surge of smart glasses~\cite{aria23surreal, vuzix, rayban, quest, hololens, avp, magicleap, vive} and other wearables. These devices not only provide seamless access to LLM-based AI assistants, but also are multimodal data-capture vehicles that provide immediate and long-term personalized context, allowing AI assistants to evolve into the next generation of human-centric contextualized AI, and unlock a new era of \textit{contextualized computing} combined with AR/VR technology. 

In this paradigm, the wearer's own body motion and action provides important context. The problem is challenging to solve given insufficient self-observations from wearable devices. In reality, a critical limiting factor to advance research is data. Currently, people either use real data with limited scale, diversity and modality~\cite{amass19iccv, hps21cvpr,egobody22eccv, adt23iccv, divatrack24eg, parahome24, egoexo4d24cvpr,truman24cvpr} or simulations that lack realism and completeness~\cite{selfpose20pami, avatarposer22eccv, tip22siggraph, unrealego22eccv, circle23cvpr, egogen24li}. There are three key technical challenges to create large motion datasets. 

\begin{itemize}
    \item{\it{Obtaining long-term ground-truth motion in the wild.}} There are two main motion capture (mocap) approaches. Vision-based solutions such as the ones relying on optical markers~\cite{amass19iccv, adt23iccv, parahome24} or cameras~\cite{totalcapture2017mbvc, egobody22eccv, egohuman23iccv, motionx23neurips, egoexo4d24cvpr} are adversely affected by line-of-sight visibility, and require a complex multi-camera-setup to cover limited range of motions inside a constrained volume. Inertial-based solutions~\cite{mvnlink,rokoko} suffer from dead-reckoning and thus are inferior in global positioning accuracy~\cite{xsens09, 3dpw18eccv}.    
    \item{\it{Multi-device alignment.}} Combining multiple capture devices or ground-truth systems requires accurate temporal and spatial alignment, which can be challenging with off-the-shelf hardware that cannot be modified or lack support for universal synchronization protocols. The existing datasets work around this problem using visual cues~\cite{egohuman23iccv,egoexo4d24cvpr, egobody22eccv} or audio~\cite{hps21cvpr}. Such approaches offer limited accuracy and reliability. In order to counter clock drift for long recordings, they can be intrusive and interrupt the natural activity. Consequently, existing datasets mostly record short motions (\cf \cref{tab:cmp}).
    \item{\it{Data processing and annotations.}} These are critical for a dataset to develop its full potential. In addition to the body motion, device localization and scene representation, we believe natural language descriptions are crucial for future research directions. Existing work provides simple descriptions or action labels without scene context~\cite{kitml16, humanml3d22cvpr, posescript22eccv}, and is of significantly smaller scale compared to the text corpus for training LLMs~\cite{gpt3, gpt4, llama}.
\end{itemize}

\begin{table}[b]
    \centering
    \setlength{\tabcolsep}{2.9pt}
    \footnotesize
    \begin{tabular}{*{13}{c}}
    \toprule
    Seq &Qty  &Pts   &Sce &Loc     &Pose  &Img  &IMU  &Gaze  &Traj     &Sent &Word &Voc \\
    \hline
    1200 &300h &\ppcnt &20 &\loccnt &260M  &201M   &11.7B &10.8M &399Km  &\sentcnt &\wordcnt &\vocsz \\
    \bottomrule
    \end{tabular}
    \caption{\textbf{Highlight statistics of \nymeria dataset}. We capture 1200 sequences of 300-hour daily activity from \ppcnt people performing \scriptcnt scenarios at \loccnt locations with 399Km traveling distance, \posecnt body poses, \imgcnt images, \imucnt IMUs, \gazecnt gazes. The motion narrations contain \sentcnt sentences in \wordcnt words from 6545 vocabulary.}
    \label{tab:stats}
\end{table}

To fill the gap and accelerate the research, we introduce \nymeria - the world largest human motion dataset with \hourcnt hours in-the-wild daily activities from \ppcnt participants performing \scriptcnt scenarios from \loccnt indoor and outdoor locations. With avarege 15-min per recording, the data captures natural activities with spontaneous unscripted actions and authentic interactions. Nymeria is first-of-its-kind dataset recorded with multiple multimodal egocentric devices. Participants worn XSens mocap suit~\cite{mvnlink}, Project Aria glasses~\cite{aria23surreal} and Aria-alike wristbands to record egocentric motion, RGB, grayscale, eye tracking (ET) videos, inertial measurement units (IMUs), magnetometer, barometer and etc. Devices are synchronized with a non-intrusive hardware solution with sub-millisecond accuracy and localized into a single metric 3D leveraging Project Aria Machine Perception Service (MPS)~\cite{mps}. We also developed novel algorithms to retarget XSens skeleton motion into a full parametric human model and correct the global drift with optimization. To connect human motion with natural languages, we developed a coarse-to-fine narration schema to describe in-context human motion at different granularity. With \sentcnt sentences and \wordcnt words from \vocsz vocabulary size, \nymeria stands out as the world's largest motion-language dataset.

\begin{table}[!hb]
    \centering
    \setlength{\tabcolsep}{1pt}
    \newcolumntype{C}[1]{>{\centering\arraybackslash}p{#1}}
    \newcolumntype{L}[1]{>{\raggedright\arraybackslash}p{#1}}
    \newcolumntype{R}[1]{>{\raggedleft\arraybackslash}p{#1}}
    \newcommand{\nan}{-}
    \footnotesize 
    \begin{tabular}{L{22mm} *{4}{R{6.8mm}} *{2}{C{8.1mm}} *{10}{R{4.5mm}}} 
    \toprule
    dataset &q/h  &p/M  &$\mu$/m &pp    &tt/K  &voc    &pm &hd       &3p      &wd        &gp      &od       &gz       &sr       &mp        &hh       \\ \midrule
    \rowcolor{lightgray!30}
    AMASS\cite{amass19iccv}          &42   &0.9  &0.22    &346   && &\checkmark  &  &  &  &  &  &  & & & \\
    HPS\cite{hps21cvpr}              &4.5  &0.5  &8.2     &7     && &\checkmark  &\checkmark &            &            &  \checkmark   & \checkmark &           &\checkmark  &\checkmark &\checkmark \\
    \rowcolor{lightgray!30}
    EgoBody\cite{egobody22eccv}      &2    &0.4  &1       &36    && &\checkmark  & \checkmark & \checkmark &            &  &            & \checkmark & \checkmark & \checkmark & \checkmark \\
    HML3D\cite{humanml3d22cvpr}      &28.6 &2.9  &0.12    &\nan  &45.0 &5371 &\checkmark  & & & &  & & & & & \\
    \rowcolor{lightgray!30}
    EgoHuman\cite{egohuman23iccv}    &3.5  &0.4 &0.5     &7      & & &\checkmark  & \checkmark & \checkmark & &  & \checkmark &           &           & \checkmark  & \checkmark \\ 
    MotionX\cite{motionx23neurips}   &144  &15.6   &0.11  &\nan  &81.1 &\nan &\checkmark  & & \checkmark & &  & \checkmark & & & & \\
    \rowcolor{lightgray!30}
    DivaTrack\cite{divatrack24eg}    &16.5 &3.6    &0.13  &22    && &  & &  &  &  &  &  &  &  &\\
    EgoExo4D\cite{egoexo4d24cvpr}    &88.8 &9.6   &\underline{2.6} &\underline{740} &\underline{432} &\underline{4405} & & \checkmark & \checkmark & & & \checkmark & \checkmark & \checkmark & & \\
    \rowcolor{lightgray!30}
    ParaHome\cite{parahome24}  &7.33   &56  &4.4  &30   & & &\checkmark  & &\checkmark & &\checkmark &  & &\checkmark & & \\ 
    LaHuman\cite{laserhuman24} &3  &\nan &0.51    &\nan &12.3 &\nan &\checkmark & &\checkmark & &\checkmark &\checkmark & &\checkmark &\checkmark &\checkmark \\ 
    \midrule
    Nymeria(ours)   &300    &260     &15 &\ppcnt   &310.5 &\vocsz & \checkmark & \checkmark & \checkmark & \checkmark & \checkmark & \checkmark & \checkmark & \checkmark & \checkmark & \checkmark \\
    \bottomrule
    \end{tabular}
    \caption{\textbf{Human motion datasets by releasing date}. Columns 2 to 5 show
    activity in hour (q/h), 
    pose frames in millions (p/M),
    mean sequence duration in minute ($\mu$/m) and
    number of participants (pp).
    We then compare language narrations \wrt number of descriptions (tt/K) and vocabulary size (voc).
    The remaining columns mark following features:
    parametric model for motion representation (pm), 
    egocentric head-mounted device (hd),
    third-person perspectives (3p),
    wristbands (wd),
    global positioning (gp),
    outdoor scene (od),
    gaze (gz),
    3D scene representations (sr),
    multi-people scenarios (mp) and
    human-human interactions (hh).
    Note EgoExo4D~\cite{egoexo4d24cvpr} reports 1422h by summing per camera recording time, where the total activity is 180h, with 88.8h annotated with MSCOCO keypoints. The underline numbers are reported for the full dataset.}
    \label{tab:cmp}
\end{table}

\section{Related Works}

\paragraph{Motion datasets -- scale, multimodal, in-the-wild and perspectives.}
Datasets are crucial ingredients in developing algorithms, particularly machine learning approaches. 
AMASS~\cite{amass19iccv} is a pioneering effort in large motion dataset, which unifies multiple marker-based datasets into SMPL~\cite{loper15smpl}. While AMASS provides diverse human motion, it lacks scene context. 
Recent works~\cite{adt23iccv, parahome24, truman24cvpr} extend the solution to include objects.
Monocular~\cite{densepose2018,smpl-x19cvpr,kanazawa2019learning, physcap20tog, frankmocap21iccv, goel2023humans, ye2023slahmr,diogo23eg,cai2023smplerx,motionx23neurips} and multi-view cameras~\cite{panoptic15iccv, panoptic17pami,egobody22eccv,egohuman23iccv,egoexo4d24cvpr} are common mocap alternatives. For monocular camera, Motion-X~\cite{motionx23neurips} stands out as a comprehensive large collection of whole-body motion with facial expressions and hand gestures. For multi-view setting, EgoExo4D~\cite{egoexo4d24cvpr} stands out as a large dataset of skilled activities. Vision-based algorithms require good line-of-sight. Consequently, they are better suited to record motion with clear body observations bounded by a volume. To capture data in the wild, mocap suit is a popular candidate~\cite{totalcapture2017mbvc, 3dpw18eccv, empose21iccv, hps21cvpr, emdb23iccv,mocapee24lee, divatrack24eg,parahome24,laserhuman24}. To address dead-reckoning for inertial-based tracking, previous works fuse IMUs with vision~\cite{totalcapture2017mbvc,3dpw18eccv,egolocate23tog}, optimize motion with 3D scenes~\cite{hps21cvpr,mocapee24lee} and limit the range of motion locally~\cite{divatrack24eg}. Simulation as a more scalable solution, offers valuable supplement to real data. Existing works leverage gaming engine for character animation~\cite{cai2021playing}, render marker-based mocap with virtual characters and scenes~\cite{unrealego22eccv, bedlam23cvpr, truman24cvpr}, simulate motion with VR~\cite{circle23cvpr} or by generative algorithms~\cite{egogen24li} etc. Simulations often struggle to present noise characteristics of real data, result in domain gaps. 
While many solutions are developed for third-person views, egocentric motion datasets remain sparse, leaving a gap to the recent advance of egocentric perception~\cite{ego4d22cvpr,hololens, aria23surreal}. Existing works focus on hands with object-interaction~\cite{epickitchen18eccv, epickitchen21pami, h2o21iccv, assembly22cvpr,hot3d24arxiv}, lack ground-truth~\cite{ego4d22cvpr} or parametric body motion~\cite{egoexo4d24cvpr}, or is limited in scale, diversity, and modality~\cite{hps21cvpr,unrealego22eccv,egopw22cvpr,egobody22eccv,egogta23cvpr,egohuman23iccv,adt23iccv, mocapee24lee}. Nymeria is designed to fill the gaps with significant delta to existing datasets (\cf \cref{tab:cmp}).

\paragraph{Motion with natural language.}
Adding language descriptions to motion leads to unique perspective in motion understanding, especially given the powerful capability of large language models (LLMs).
Early datasets~\cite{ionescu2013human3, guo2020action2motion, ghorbani2021movi, harvey2020robust, shahroudy2016ntu, cai2022humman} offer sparse annotations as action categories or semantic labels~\cite{babel21cvpr}. KIT~\cite{kitml16} is the first attempt in using complete sentences to describe locomotion. HumanML3D~\cite{humanml3d22cvpr} enriches HumanAct12~\cite{guo2020action2motion} and AMASS~\cite{amass19iccv} with multiple descriptions per motion clip. Motion-X~\cite{motionx23neurips} leverages narration algorithm to obtain large-scale fine-grained descriptions at sequence and frame levels. Compared to available text corpus, motion-language data is rather sparse with brief text on brief motion~\cite{laserhuman24}. Ego4D~\cite{ego4d22cvpr} and EgoExo4D~\cite{egoexo4d24cvpr} offer large-scale descriptions of atomic actions, however, ground-truth motion is missing or without parametric representations. In our work, we provide in-context coarse-to-fine narration by annotators, with the amount of data a magnitude larger than prior works.

\paragraph{Egocentric body tracking, synthesis and action recognition.}
Tracking one's own body motion with wearable devices is challenging. Early approaches leverage dense body-worn IMUs~\cite{sip17eg, dip18tog, 3dpw18eccv}. Recent methods reduce IMUs to improve practicality~\cite{transpose21tog,tip22siggraph, lobstr21cgf, pip22cvpr}, assume IMU-enabled AR/VR devices~\cite{avatarposer22eccv, egoposer23}, combine IMU with cameras~\cite{hybridfusion18eccv, hps21cvpr, egopw22cvpr, egogta23cvpr, egolocate23tog}, mobile phones~\cite{imuposer23chi} and wristbands~\cite{mocapee24lee}. While sparse sensors are more practical, our work centers on constructing an offline dataset to serve ``ground truth''. Consequently, we use dense IMUs to ensure high accuracy and correct global drift via optimization. Since full-body motion is ill-posed given insufficient observation from egocentric headset, motion synthesis is often used to produce plausible motion. To condition the generation, research explore sparse motion measures~\cite{divatrack24eg, bodiffusion2023iccv, diffuseposer24cvpr}, headset motion~\cite{egoego23cvpr} and eye gaze~\cite{gimo22eccv} and text~\cite{temos22eccv, guo2022tm2t, motionclip22eccv}. The success of diffusion models lead to active develop in text-driven motion synthesis~\cite{dabral2022mofusion, gmd23iccv, mdm23iclr, priormdm24iclr, tedi23, mld23cvpr}. Similarly LLMs inspire novel motion understanding algorithms~\cite{lucas2022posegpt, motiongpt23arxiv, motiongpt24neurips, chatpose24cvpr} that tightly couple motion with natural language. Our work constructs a dataset with rich hierarchical narrations to inspire further breakthroughs in the field. 
\section{Building Nymeria Dataset}
    \input{fig_hw}
    \subsection{Data collection setup}
    \subsubsection{Hardware.}

    Each participant wears a mocap suit, a pair of glasses, two wristbands, and a synchronization device (\cf \cref{fig:capture-setup}. 
    XSens MVN Link~\cite{mvnlink} is adopted for mocap, which is a tight-fit body suit wired with 17 inertial trackers and a magnetometer. MVN Link supports on-device recording, making it ideal to collect in-the-wild data. We use Project Aria glasses~\cite{aria23surreal} as a lightweight headset to record multimodal data. The sensor suite includes 1 RGB camera, 2 grayscale peripheral cameras, 2 ET cameras, 2 IMUs, 1 barometer, 1 magnetometer, 7 microphones, 1 thermometer, GNSS, WiFi and BT. We repackaged the electronics and sensors of Project Aria into a new wristband device called \textit{miniAria}, in order to closely resemble current AR/VR headsets and provide data to better constrain body tracking algorithms. The wristband matches Project Aria's sensing ability, with exclusion of microphones, barometer, magnetometer and ET cameras. The supplementary provides detailed sensor configuration and recording profiles. 
        
    \subsubsection{Synchronization.} 
    Project Aria can record an externally provided time signal to aid synchronization. We further enable MVN Link to accept the same signal. A synchronization device is developed to supply the timestamps for all devices, which can optionally receive time from a wireless server located in radio range ($\sim$100m). This facilitates synchronizing multiple devices with sub-millisecond accuracy. The alignment between XSens and Aria is within 1 motion frame \ie 4.2~ms. To capture multiple people simultaneously, we replicate the described setup per participant and leverage a common time server.  

    \input{fig_dataset}
    \subsubsection{Recording protocols.} 
    Data collection is managed by 2-3 onsite operators. In addition to participants, a trained observer wearing Project Aria is present to record participants from third-person perspective. All people interact naturally as per activity requires, contributing to rich dynamics as opposed to staged motion. To complete each recording, participants first perform a brief mocap calibration, then gaze calibration, and finally 15-20min activity. We collect 4-8 recordings per person, where a bulk of data is captured at family houses.
            
    \subsubsection{Scenarios.}
    We define \scriptcnt scenarios (\cf examples in \cref{fig:teaser} and \cref{fig:diversity}). For indoor activities, scenarios include cooking, working, entertaining, searching objects etc. For outdoor activities, scenarios include hiking, biking, dining, sports etc. To encourage natural interactions and authentic motions, participants are instructed with high-level guidelines \eg ``grab food in the cafeteria and eat on the patio''. Operators also prompt in-context actions to increase dynamics.
    
    \subsubsection{Privacy considerations.}
    We follow Project Aria research guideline for responsible innovation. Prior to data collection, consents were obtained from participants and home owners for recording and data usage. The SOTA de-identification algorithm EgoBlur~\cite{egoblur} is used to blur faces and license plates for all videos.

\subsection{Data processing}
To process data, we first use XSens software to obtain the skeleton motion, and Project Aria MPS~\cite{mps} for device localization, scene representation and gaze estimation. Then motion is retarget to a parametric human model~\cite{momentum} and registered into the coordinates of Aria devices via optimization. 

\subsubsection{Full-body mocap and retargetting.} 
    We record motion at 240Hz, following the recommended procedures: 1) carefully measuring body dimensions of participants;
    2) performing calibration prior to every recording; and 3) processing XSens with the highest quality with single- or multi-floor specification. 
    
    XSens represents the body motion as the global transformation and 3D local joint angles of a template skeleton. The skeleton consists of 23 segments, where each segment matches the measured body dimensions of the subject. In addition, a set of $K=79$ anatomical landmarks are defined on the skeleton model~\cite{mvnmanual}. Their global positions, $\{\mathbf{p}_i\}_K$, can be computed by evaluating the forward kinematics at each frame. We utilize these landmarks to retarget the body motion onto an anatomically-inspired human model for improved realism and visual validation. Our human model is parameterized by $\{\boldsymbol{\theta}, \boldsymbol{\phi}\}$, where the pose parameters $\boldsymbol{\theta}$ define the global transformation and local joint angles, and the shape parameters $\boldsymbol{\phi}$ represent a global body scale and individual bone length. Given a motion of $N$ frames, we solve the following inverse kinematics optimization problem: 
    \begin{equation}
    \underset{\boldsymbol{\phi}, \boldsymbol{\theta}_{0\cdots T-1}, \mathbf{v}^{0\cdots K-1}}{\arg\min}
    \sum_{t=0}^{N-1}\sum_{i=0}^{K-1}\left\|T^{i}(\boldsymbol{\phi}, \boldsymbol{\theta}_t)\mathbf{v}^i - \mathbf{p}_t^{i} \right\|^2 \;,
    \end{equation}
    where $\mathbf{v}^i$ is the local offset of the $i$th landmark defined on our model, and $T^i$ is the global transformation of its parent joint. We initialize $\mathbf{v}^i$ by manually placing them on the model. The supplementary provides more details about our human model and motion retargetting.
    
     \begin{figure}[tb]
        \centering
        {
        \setlength{\fboxsep}{0pt}\setlength{\fboxrule}{0.5pt}
        \begin{minipage}{0.517\linewidth}%
        \fbox{\includegraphics[width=\linewidth]{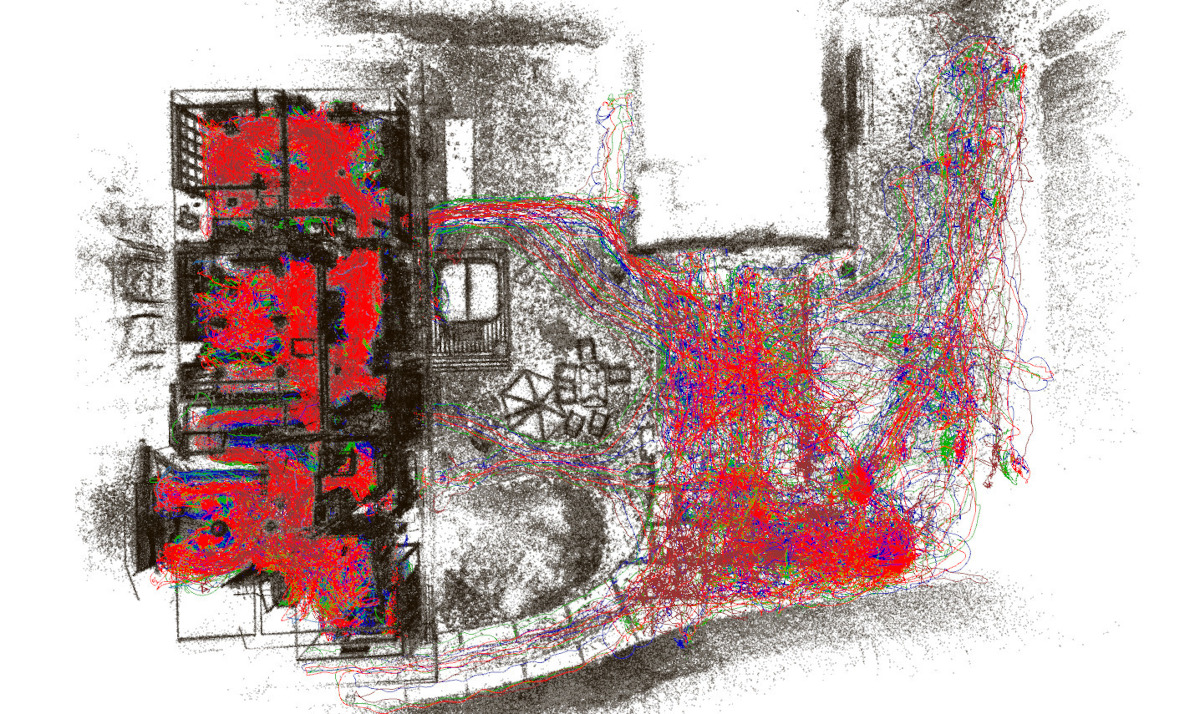}}%
        \end{minipage}~%
        \begin{minipage}{0.15\linewidth}%
        \fbox{\includegraphics[width=\linewidth]{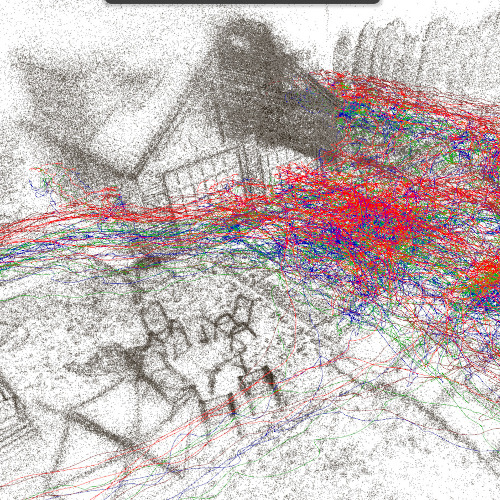}}\\[0.05mm]%
        \fbox{\includegraphics[width=\linewidth]{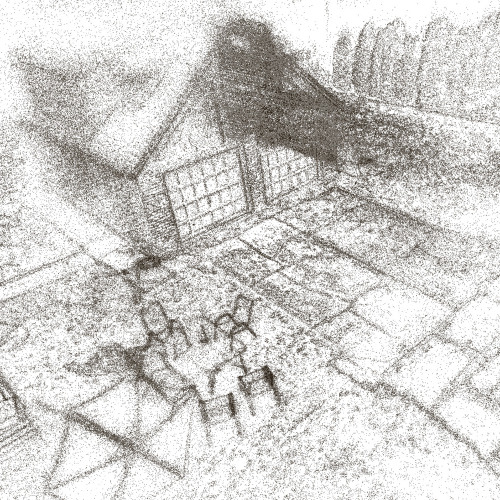}}%
        \end{minipage}~%
        \begin{minipage}{0.15\linewidth}%
        \fbox{\includegraphics[width=\linewidth]{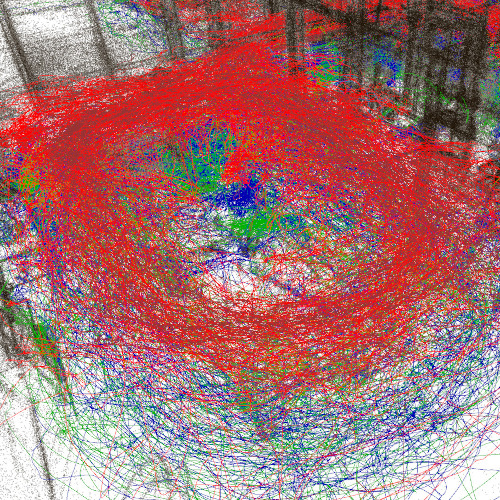}}\\[0.05mm]%
        \fbox{\includegraphics[width=\linewidth]{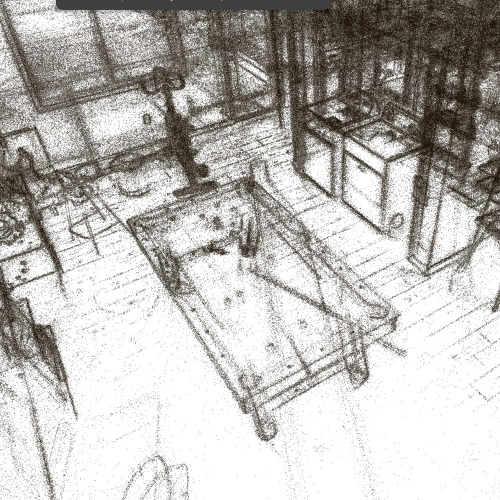}}%
        \end{minipage}~%
        \begin{minipage}{0.15\linewidth}%
        \fbox{\includegraphics[width=\linewidth]{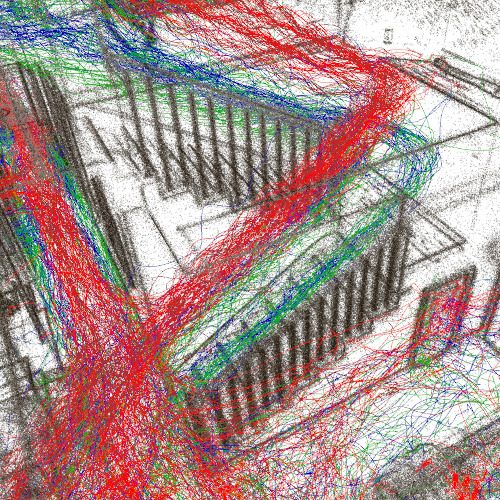}}\\[0.05mm]%
        \fbox{\includegraphics[width=\linewidth]{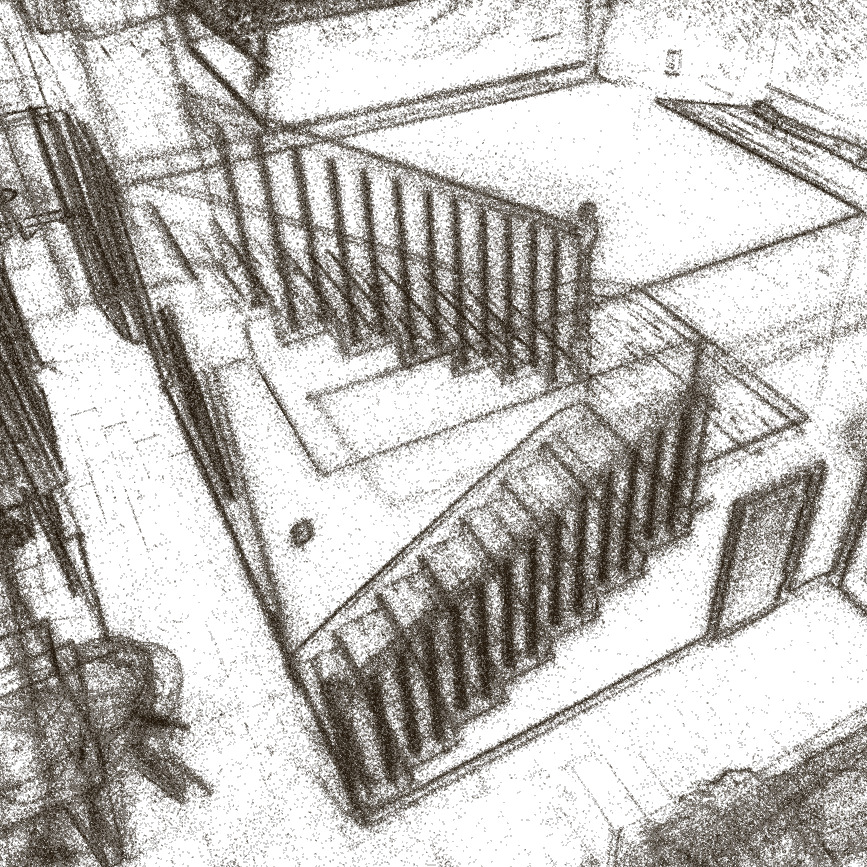}}%
        \end{minipage}~%
        }\\
        {
        \setlength{\fboxsep}{0pt}\setlength{\fboxrule}{0.5pt}
        \begin{minipage}{0.517\linewidth}%
        \fbox{\includegraphics[width=\linewidth]{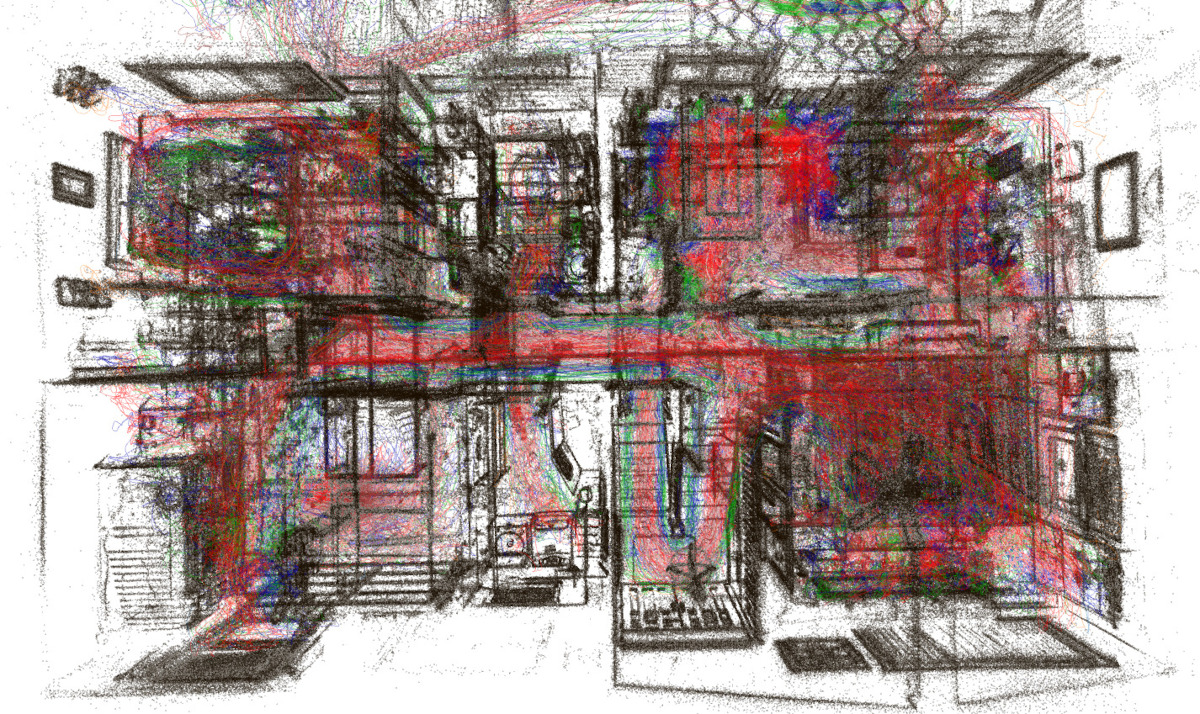}}%
        \end{minipage}~%
        \begin{minipage}{0.15\linewidth}%
        \fbox{\includegraphics[width=\linewidth]{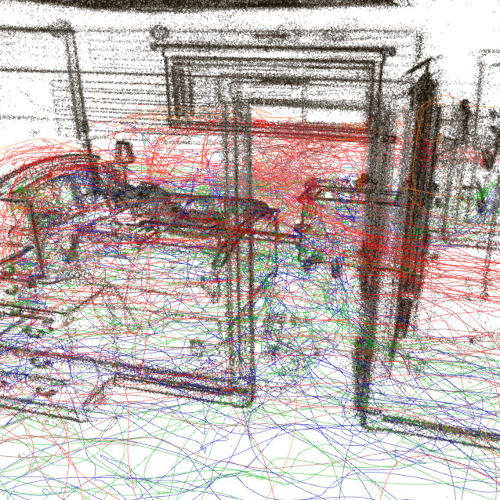}}\\[0.05mm]%
        \fbox{\includegraphics[width=\linewidth]{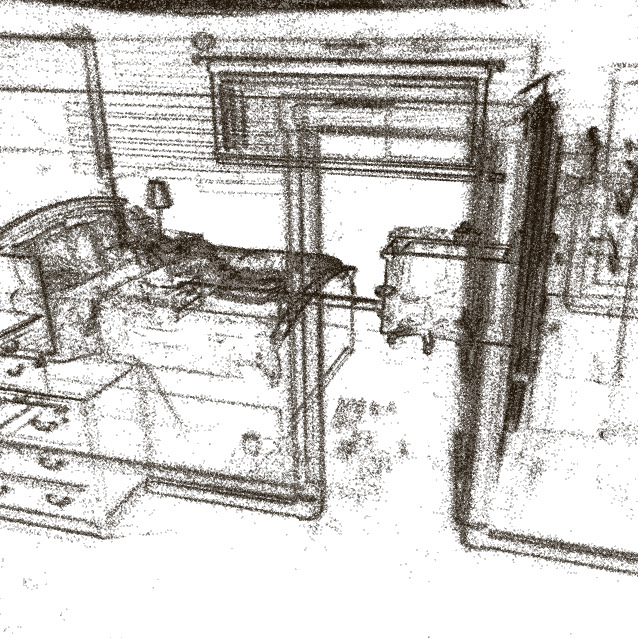}}%
        \end{minipage}~%
        \begin{minipage}{0.15\linewidth}%
        \fbox{\includegraphics[width=\linewidth]{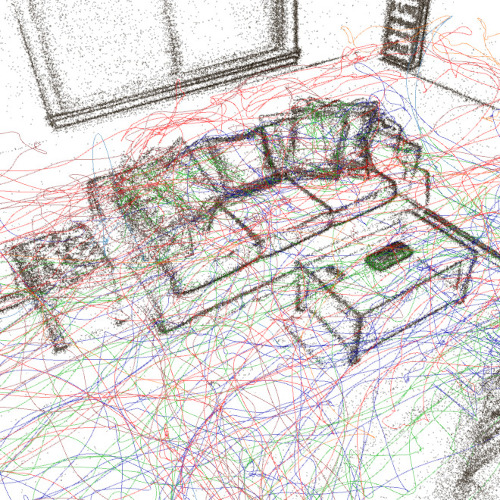}}\\[0.05mm]%
        \fbox{\includegraphics[width=\linewidth]{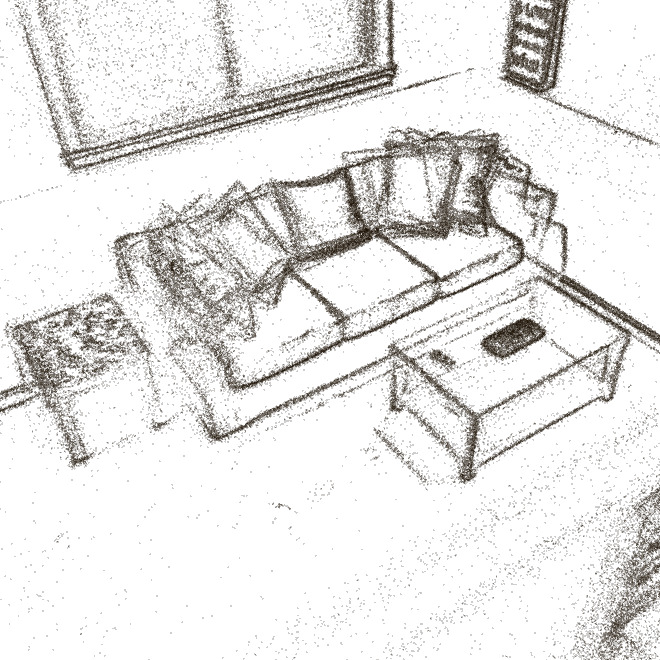}}%
        \end{minipage}~%
        \begin{minipage}{0.15\linewidth}%
        \fbox{\includegraphics[width=\linewidth]{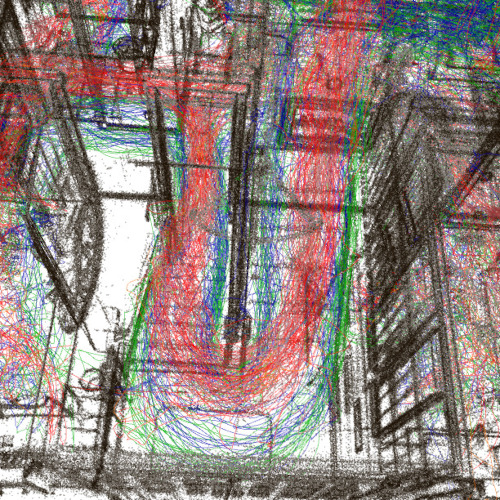}}\\[0.05mm]%
        \fbox{\includegraphics[width=\linewidth]{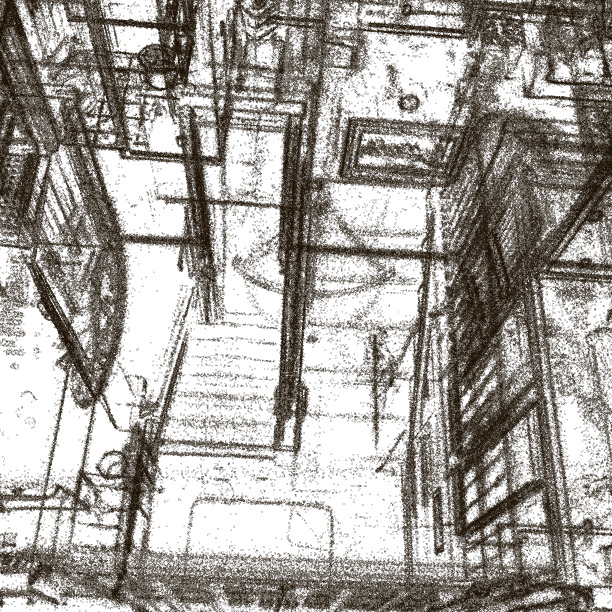}}%
        \end{minipage}~%
        }\\[0.1mm]
        \caption{\textbf{Global aligned trajectories and point clouds by locations}. We show examples of split-level residential house with gardens, where each contain $\approx$5 hours of recording. The left shows the top-down views of accumulated trajectories where red, green and blue indicate the head, the left and right wrist. On the right we sample closed-up views where the clusters emerge from human 3D motion distribution.} 
        \label{fig:airbnb}
     \end{figure}
     
\subsubsection{6DoF localization and mapping with global alignment.}
     Data recorded at the same location are globally aligned into a single metric 3D world via Project Aria MPS~\cite{mps}, which employs state-of-the-art visual inertial odometry (VIO), SLAM and mapping algorithms~\cite{MSCKF, orbslam, dso}. In a nutshell, first SLAM is run for each individual recording independently. Subsequently, the resulting maps are loop-closed and jointly optimized via visual-inertial bundle adjustment. The output are highly accurate 1KHz trajectories (\cf \cref{fig:airbnb} and supplementary) -- allowing for example to visualize head- and wrist-motion-clusters respectively.
         
     Given the device trajectories, we align the body poses into this same reference coordinates by correlating the dead-reckoned trajectory from XSens. Since the latter accumulates significant drift, there exists no static global transformation to align them. Instead, we assume a constant transformation $T_{HD}$ between the user's head segment from XSens $H$ and the Aria device $D$ (Aria is firmly held in place with straps and participants are asked to avoid adjusting the glasses during the recording). We then cut the trajectory into a large number of short 4.2\,ms segments, and solve the following optimization problem
     \begin{equation}\label{eq:handeye}
     \underset{T_{HD}}{\arg\min} \sum_t \left\|\log\left(\left( {T_{OH}^t}^{-1} {T_{OH}^{t+1}} \right) \cdot\left( T_{HD} {T_{WD}^t}^{-1}  T_{WD}^{t+1} {T_{HD}}^{-1} \right)^{-1} \right)\right\|^2,
     \end{equation}
     where $O$ is the drifting odometry frame of XSens and $W$ is the world coordinates of the MPS output. The is a HandEye calibration problem with closed-form solution~\cite{sorkine17:svd}. Note the formulation effectively aligns a large number of local motion clips by comparing the local velocity.
     In practice, Aria is not completely rigid during recording, resulting in a main source of inaccuracy. Precision can be improved with a rolling window optimization. \Cref{fig:projection} provides a qualitative end-to-end assessment of our multi-device location, XSens motion registration, and time synchronization.

    \input{fig_projection}
    
\subsection{In-context motion-language description}
    \input{fig_eg_narration}

    \input{fig_wordcloud}
    To build the connection between body motion, natural language and activity recognition, we ask human annotators to write textual descriptions of in-context human motion by viewing playback videos of the dataset. The annotators segment the video into clips to write descriptions by answering predefined questions. To give annotators a holistic understanding of the motion, the playback video contains synchronized views of the egocentric video, third-person video, and human motion rendered with 3D scene.
    
    We define three annotation tasks to describe motion coarse to fine, and scale up human efforts in a meaningful way. The finest level is \textit{motion narration} for detailed body posture, \eg motion direction, velocity, interactions and attention. Next, we annotate for \textit{atomic action}. Compared to motion narration, annotators are encouraged to use verbs whenever possible, \eg using ``dancing'' instead of ``swing both arms while rotating the body to the right with legs slightly apart''. The two tasks are done for <5s clips. The last task \textit{activity summarization} is to give one-sentence summary over 30s activity. \Cref{fig:narration-examples} provide examples of each annotation task. The figure shows the benefit of providing annotators three synchronized views. While the egocentric view captures closed-up hand-object interactions, the third-person and motion rendering help annotators grasp a holistic understanding of the actions.

\begin{figure}[tb]
    \centering
    \includegraphics[width=1\textwidth]{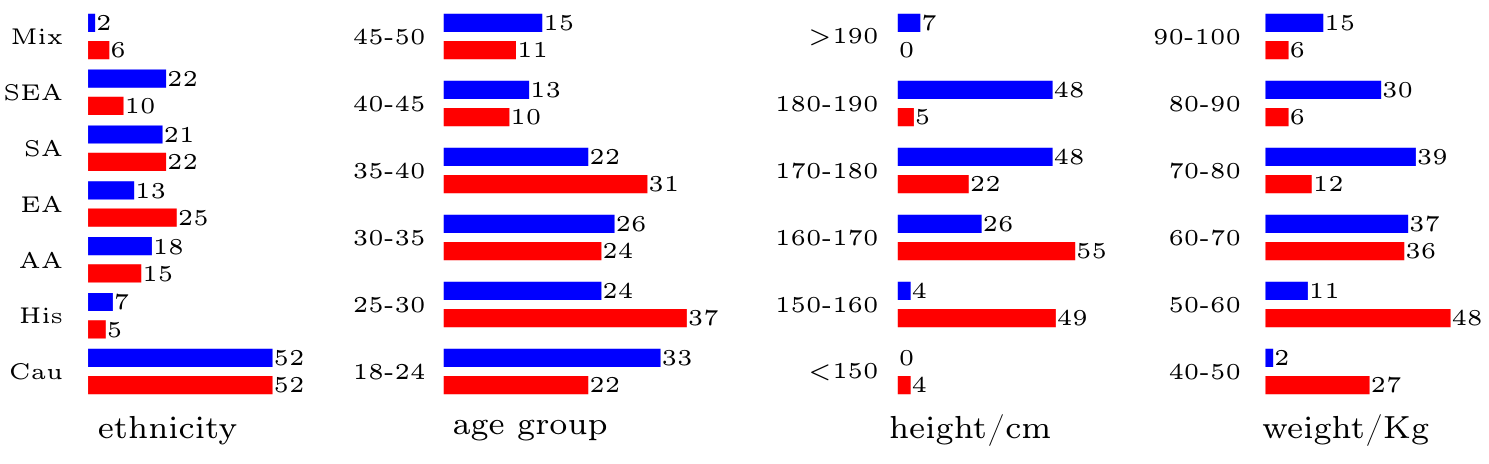}
    \caption{\textbf{Demographics by \textcolor{red}{female} and \textcolor{blue}{male}.} Meaning of symbols: Caucasian (Cau), Hispanic (His), African American (AA), East/South/Southeast Asian (EA/SA/SEA).}
    \label{fig:demog}
\end{figure}
\subsection{Statistics}
    \paragraph{Data statistics.}
    We collected \hourcnt daily activities from \ppcnt participants, which amounts to \seqcnt sequences with average 15min duration. The accumulated trajectory from all participants is \headtl for headset and \wristtl for both wristbands. \Cref{fig:demog} shows the participant demographics \wrt the self-reported ethnicity, age, height and weight. The statistics is split by gender, where 48.5\% participants self-identified as female, and 51.4\% as male. The dataset captures \abcnt houses, where \multifloor are multi-floor. In total, there are \rooms rooms and \gardens gardens. We also capture three locations from an open campus, including 1 cafeteria with an outdoor patio, 1 office building, and a public parking connected to multiple biking/hiking trails. 
    Approximately 15\% recordings contain outdoor activities. Among all scenarios, the highest occurrences are cooking, searching objects and improvised actions. We provide detailed breakdown in supplementary.
    \paragraph{Annotation statistics.}
    For language descriptions, we annotated \motioncnt motion narration, \atomiccnt atomic action and \activitycnt activity summarization. The average video segment is 5 seconds for narration and atomic action and 30 seconds for summarization. In total, the dataset provides \sentcnt sentences \wordcnt words from a vocabulary of 6545 distinctive words. Note the average word per sentence is 27.8, which is significant longer than existing motion-language narrations. \Cref{fig:wordcloud} visualizes the language distribution.

\section{Benchmark Tasks and Baselines}\label{sect:exp}
\subsection{Research opportunities}
With an enormous amount of contextualized human motion, Nymeria dataste provides unprecedented research opportunities. Following, we highlight a few research domains. This is by no means complete, but to inspire novel directions and ideas by scratching the surface of full potential.

\paragraph{Motion tasks.} 
Nymeria is created to assist human motion understanding, with an emphasis on, but not limited to, egocentric perception. The data supports various topics, \eg full-body tracking, motion synthesis, motion forecasting, path planning, action recognition, human behavior analysis and etc. With a multimodal multi-device dataset, we encourage exploring unique algorithms with novel settings, \eg gaze-conditioned motion prediction, action recognition from headset and wristbands, interaction generation from language etc. 

\paragraph{Multimodal spatial reasoning and video understanding.} In addition to body motion, \nymeria also provides extensive egocentric videos with precise localization and synchronization. \nymeria therefore is a great asset for algorithms requiring camera poses as priori, \eg scene reconstruction~\cite{nerf20eccv, gs23siggraph}. In this regard, the rich dynamics raise novel real-world challenges. Nymeria also facilitates video understanding with numerous in-context narration. Image retrieval and relocalization also benefit from \nymeria, sincell we capture multiple videos per location and align them in global coordinates.

\paragraph{Simulation.} Simulation is a promising technique to gather massive data, however, synthetic data is typically seed in real world~\cite{bedlam23cvpr, truman24cvpr}. By design, Nymeria is naturally useful to drive in-context character animation. Since human motion is a function of the environment, we believe Nymeria also aid simulating 3D scene~\cite{mime23cvpr}. Sensor simulation can benefit from our data as well, especially for IMUs, magnetormeter, and barometer in combination with motion priors.

\begin{table}[t]
\setlength{\tabcolsep}{2.2pt}
\renewcommand{\arraystretch}{1.1}
    \begin{minipage}[t]{0.5\textwidth}
    \centering
    \scriptsize
    \begin{tabular}{l|ccc|c|c}
    \toprule
    & \multicolumn{3}{|c|}{MPJPE (cm)} & Hand & MPJVE \\ 
    \cline{2-4}
    & Mean & Lower & Upper & PE(cm) & (cm/s)\\  
    \midrule
    AMASS &4.20 &8.06 &1.88 &2.34 &28.23 \\ 
    Real & 7.97 & 16.74 & 3.13 & 6.25 & 16.71 \\
    Synthetic & 7.31 & 15.97 & 2.51 & 3.47 & 16.63 \\
    \bottomrule
    \end{tabular}
    \vspace{5pt}
    \caption{AvatarPoser~\cite{avatarposer22eccv} train/test with real vs. synthetic poses from Nymeria. The first row reports AvatarPoser train/test on AMASS~\cite{amass19iccv} for reference.}
    \label{tab:ap}
    \end{minipage}
\hskip1em
    \begin{minipage}[t]{0.45\textwidth}
    \centering
    \scriptsize
    \begin{tabular}{l|ccc|c}
    \toprule
    & \multicolumn{3}{c|}{MPJPE (cm)} & \multirow{2}{*}{FID} \\
    \cline{2-4}
    & Mean & Lower & Upper & \\ 
    \midrule
    BoDiffusion(A) &3.63 &7.07 &1.53 & - \\
    BoDiffusion & 7.98 & 15.27 &5.28 & 2.32 \\
    EgoEgo & 13.22 & 19.03 & 10.00 & 5.14 \\
    \bottomrule
    \end{tabular}
    \vspace{5pt}
    \caption{Motion synthesis with EgoEgo~\cite{egoego23cvpr} and BoDiffusion~\cite{bodiffusion2023iccv} on Nymeria. BoDiffusion(A) reports results train/test on AMASS~\cite{amass19iccv}.}
    \label{tab:diffusion}
    \end{minipage}
\end{table}

\subsection{Motion tasks baselines}
We use three case studies to showcase motion algorithms on Nymeria. The goal is to validate the data and provide future research with common baselines. Due to page limit, details of model training is present in supplementary.

\subsubsection{Motion tracking and synthesis from sparse inputs.} 
In this case study, we take the 1-point/3-point body tracking problem in modern AR/VR, where people wear a headset and optionally controllers in VR, and a pair of glasses and optionally wristbands in AR. The task is to recover wearer's full-body motion using only headset (1-point)~\cite{kinpoly2021neurips, egoego23cvpr} or additional wrist devices (3-point)~\cite{avatarposer22eccv, divatrack24eg, bodiffusion2023iccv, egoposer23, du2023avatars}. The problem is considered mixture of tracking and synthesis, due to insufficient lower-body observation. Existing works simulate sensor inputs from motion datasets, which lack the noise characteristic of real data. Nymeria is the first large dataset with multimodal real data to support model training and evaluation. In particular, we provide raw IMU and device poses, as opposed to filtered~\cite{xsens09} acceleration and velocity released in previous datasets~\cite{totalcapture2017mbvc, hps21cvpr,divatrack24eg}. 
In the first experiment, we adapt the 3-point regression method AvatarPoser~\cite{avatarposer22eccv} to train on Nymeria with two variants -- one model trained with ``real'' device poses from SLAM and the other trained with ``synthetic'' input mocked-up from body motion same as in the original work. The evaluation uses the same metric as AvatarPoser, including mean per joint position error (MPJPE), hand position error and mean per joint velocity error (MPJVE). Results are shown in \cref{tab:ap}. As expected, ``real'' data lead to worse performance due to additional error-sources. However the gap is small, which indicates the quality of device tracking. Compare to the original AvatarPoser trained on AMASS~\cite{amass19iccv}, Nymeria yields reasonable but higher MPJPE, mainly from worse lower-body tracking. We hypothesize our data contain harder motion, \eg hiking uneven terrain, taking stairs, playing sports and etc. The second experiment evaluates two diffusion models for motion synthesis, \ie 3-point Bodiffusion~\cite{bodiffusion2023iccv} and 1-point EgoEgo~\cite{egoego23cvpr}. In addition to MPJPE, we report the Frechet Inception Distance (FID) which measures the distribution distance between generated motion and real motion, following the same procedure in~\cite{action2motion20} (\cf \cref{tab:diffusion}). Results trained from our data are comparable with the original works trained on subset of AMASS, where 3-point diffusion yields better performance as expected.

\subsubsection{Representing human motion manifold.}
\begin{wraptable}{r}{0.45\textwidth}
    \centering
    \scriptsize
    \setlength{\tabcolsep}{1.8pt}
    \begin{tabular}{c|c|c|ccc}
    \toprule
    PQ & CB & Dim & \scriptsize{MPJPE} & \scriptsize{PA-MPJPE} & \scriptsize{ACC} \\
    \midrule
    * & 512  & - & 55.80 & 40.10  & 7.50 \\ 
    1 & 2048 & 512 & 51.60 & 37.55 & 1.09 \\
    2 & 2048 & 512 & 39.63 & 29.77 & 0.71 \\
    2 & 4096 & 512 & 39.20 & 29.66 & 0.82 \\
    2 & 16384 & 64 & \textbf{34.49} & \textbf{26.83} & \textbf{0.67} \\
    \bottomrule
    \end{tabular}
    \caption{Ablation of motion VQ-VAE trained on Nymeria (metric unit: mm). We compare product quantization (PQ), codebook (CB) size and latent dimensions (Dim). The first row show results of AMASS as reported in~\cite{motiongpt24neurips}. PA stands for Procrustes-aligned and ACC for joint position acceleration.}
    \label{tab:vqvae}
\end{wraptable}
Learning the embedding space to model human motion manifold has many benefits, \eg dimension reduction, learning motion priors, motion denoising by projection, motion synthesis by sampling and interpolation etc. Previous attempts mainly rely on AMASS~\cite{amass19iccv} to learn the representation~\cite{smpl-x19cvpr, tiwari22posendf, rempe2021humor, ling2020character}, where the data heavily focused on isolated locomotion or professional motion. By capturing rich daily activities of common people interacting with real world, our data distribution can better represent the motion manifold concerning everyday human activities. 
To take a stab in this direction, we to train Vector-Quantized Variational Autoencoder (VQ-VAE)~\cite{vq2017neurips, dhariwal2020jukebox} for Nymeria motion data, following the previous work~\cite{motiongpt24neurips}. VQ-VAE can be used as a ``motion tokenizer'' to generate motion with auto-regression~\cite{lucas2022posegpt}, and to de-noise motion by projecting the input onto the  manifold~\cite{rempe2021humor}. An ablation is perform to study the impact of product quantization, codebook size and latent dimension. For evaluation, we adopt the same metrics in~\cite{motiongpt24neurips} and include the VQ-VAE trained on AMASS for comparison. As shown in \cref{tab:vqvae}, Nymeria motion data can be well tokenized to achieve similar performance as VQ-VAE trained with AMASS data. By leveraging product quantization, increasing the codebook size and decreasing the latent dimension, the reconstruction quality is further improved. The resulting motion tokenizer can therefore be used with LLMs akin to language tokenizer to foster motion understanding~\cite{motiongpt23arxiv, motiongpt24neurips} as detailed in the next case study.

\subsubsection{Motion and language.}
\begin{wraptable}{r}{0.52\textwidth}
    \centering
    \scriptsize
    \begin{tabular}{l|c|c|c|c|c}
    \toprule
    & Bert & Bleu@1 & Bleu@4 & CIDEr & RougeL \\ 
    \midrule
    TM2T & 11.08 & 40.11 & 8.99 & 20.85 & 30.70 \\
    MotionGPT & \textbf{14.09} & \textbf{42.22} & \textbf{10.31} & \textbf{37.27} & \textbf{32.33} \\
    \bottomrule
    \end{tabular}
    \caption{Evaluation of motion-to-text. Models are trained with small subset of Nymeria.}
    \label{table:text}
\end{wraptable}
While parametric human motion is useful for machine algorithms, natural language description is a better interface with human. An valuable feature of Nymeria dataset is the high-quality hierarchical narrations. Compared with existing data~\cite{humanml3d22cvpr,motionx23neurips}, our narrations are in longer natural sentences with context descriptions of objects and environments. It can be used to learn models for text-driven motion generation and motion-to-text descriptions. More importantly, the contextual descriptions are not only paired with the human motion, but also with videos, point clouds, and other sensory data and annotations. The corroboration of both 2D and 3D environment information with language offers exciting opportunities in grounding language and motion research in the physical world.
Leveraging VQ-VAE experiment, we train MotionGPT~\cite{motiongpt24neurips} and TM2T~\cite{guo2022tm2t} for the motion-to-text task with a subset of 30h motion narrations from Nymeria (\cf \cref{table:text}) to be directly comparable with previous results, using the same metrics of BERT~\cite{zhang2019bertscore}, BLEU~\cite{papineni2002bleu}, CIDEr~\cite{vedantam2015cider} and ROUGE-L~\cite{lin2004rouge}. TM2T performs worse than MotionGPT since it lacks strong language prior with the T5~\cite{raffel2020exploring} backbone. Compare to the original works trained with HumanML3D~\cite{humanml3d22cvpr} and KIT~\cite{kitml16}, our results are worse as expected. Given similar hours of data, our narrations are much more complex and diverse. By using the full narration data, we expect the performance to be significantly better.

\section{Conclusions and Discussions}
We propose \nymeria dataset to accelerate research in egocentric motion understanding. The dataset is the world's largest collection of human motion in the wild with \hourcnt hours daily activity, \posecnt body poses of \ppcnt participants across \loccnt locations. We provide accurate 6DoF tracking, 3D scene points and gaze, with all modalities synchronized and aligned into one metric 3D world. Collectively, the dataset captured \headtl of travel by the participants for a total of \imgcnt egocentric images, \imucnt IMU samples and \gazecnt gaze point. The \nymeria dataset also stands out as the world largest motion-language dataset with \sentcnt sentences in \wordcnt words with \motioncnt of fine-grained motion narration, \atomiccnt atomic actions and \activitycnt activity summarization.

\textit{Limitations.} The mocap suit and wristbands lead to unnatural appearance in videos, and restrict certain range of motion. XSens quality is known to be affected by motion calibration and body measurements. Our dataset only covers a portion of daily activities, leaving out common public scenarios. 

\textit{Social impact.} Understanding egocentric full-body motion is crucial towards contextual AI, however it heavily involves personal data. We make best effort to respect privacy via consent, de-identification, minimum data retention and permissive research license.

\section*{Acknowledgements}\label{supp:ack}
We gratefully acknowledge the following colleagues for their valuable discussions and technical support.
Genesis Mendoza,
Jacob Alibadi,
Ivan Soeria-Atmadja,
Elena Shchetinina, and
Atishi Bali worked on data collection.
Yusuf Mansour supported gaze estimation on Project Aria.
Ahmed Elabbasy, 
Guru Somasundaram,
Omkar Pakhi, 
and
Nikhil Raina supported EgoBlur as the solution to anonymize video and explored bounding box annotation.
Evgeniy Oleinik, 
Maien Hamed, 
and 
Mark Schwesinger supported onboarding \nymeria dataset into Project Aria dataset explorer and data release.
Melissa Hebra helped with coordinating narration annotation. 
Edward Miller served as research program manager.
Pierre Moulon provided valuable guidance to open source code repository. 
Tassos Mourikis, 
Maurizio Monge,
David Caruso, 
Duncan Frost,
and 
Harry Lanaras provided technical support for SLAM.
Daniel DeTone, 
Dan Barnes, 
Raul Mur Artal, 
Thomas Whelan, and 
Austin Kukay provided valuable discussions on annotating semantic bounding box. 
Julian Nubert adopted the dataset for early dogfooding.
Pedro Cancel Rivera,
Gustavo Solaira, 
Yang Lou, and
Yuyang Zou provided support from Project Aria program.
Svetoslav Kolev provided frequent feedback.
Arjang Talattof supported MPS.
Gerard Pons-Moll served as senior advisor.
Carl Ren and Mingfei Yan served as senior managers. 

\paragraph{Contribution Statements.} 
Lingni Ma led the project, developed the pipeline to construct the dataset, coordinated data collection/processing, and led the baseline evaluations. 
Yuting Ye developed the solution for human motion retargetting, advised data collection and evaluations. 
Fangzhou Hong, Vladimir Guzov and Yifeng Jiang, validated the dataset and implemented baselines. 
Rowan Postyeni supported daily operations, processed XSens motion data and performed quality assessment for XSens motion and narration. 
Luis Pesqueria served the program manager for data collection and narration. 
Alexander Gamino was responsible for multi-device tracking. 
Vijay Baiyya was responsible to Project Aria MPS scaled processing.
Hyo Jin Kim supported narration annotations. 
Kevin Bailey and David Soriano Fosas lead hardware development of miniAria wristband. 
C. Karen Liu and Ziwei Liu served as technical advisor to data collection, annotation and baseline evaluations. 
Jakob Engel, Renzo De Nardi and Richard Newcombe were the senior technical and scientific advisors. 

% ---- Appendix ----
\clearpage
\appendix
\subsection*{\large Appendix}

The appendix provides further details about the dataset and algorithms. The content is structured as follows.
\Cref{app:hw} gives more details about the capture setup. 
\Cref{supp:processing} provides descriptions about data processing, including a brief summary of Project Aria machine perception service (\ref{supp:mps}), 
XSens human model retargetting (\ref{supp:blueman}),
and motion-language annotation interface (\ref{supp:narration}).
\Cref{supp:facts} provides additional information of data recording scenarios (\ref{supp:scenario}) and locations (\ref{supp:locations}).
\Cref{supp:benchmark} presents details about the baseline experiments.

\section{Hardware}\label{app:hw}
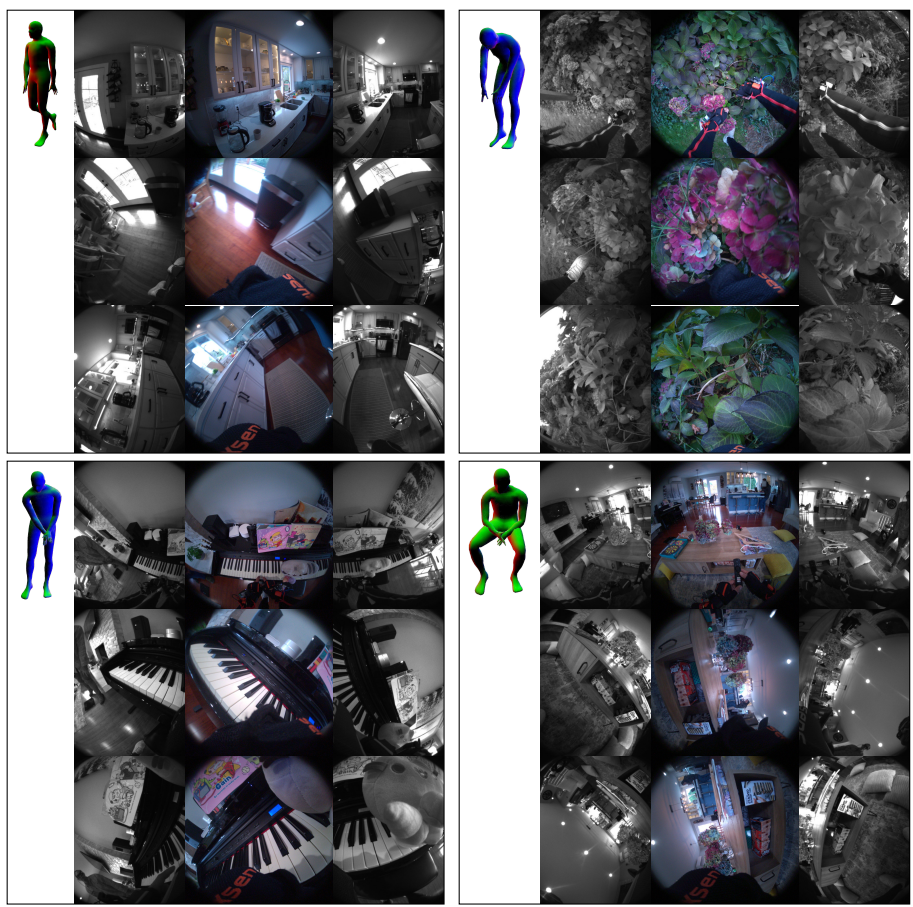
\begin{figure}
    \centering
    \begin{tikzpicture}[inner sep=0pt]
        \def\ih{1.95cm}
        \tikzset{
        pics/aria/.style n args={3}{
            code={
                \node(p0) at (#2,#3) {\includegraphics[height=\ih]{fig_vphead_#1.png}};
                \node(p1) at (p0.east)[anchor=west,xshift=-0.1em] {\includegraphics[height=\ih]{fig_vphead_#1_mono0.jpg}};
                \node(p2) at (p1.east)[anchor=west] {\includegraphics[height=\ih]{fig_vphead_#1_rgb0.jpg}};
                \node(p3) at (p2.east)[anchor=west] {\includegraphics[height=\ih]{fig_vphead_#1_mono1.jpg}};
                \node(pl1) at (p1.south)[anchor=north]{\includegraphics[height=\ih]{fig_vpwrist_#1_mono0.jpg}};
                \node(pl2) at (pl1.east)[anchor=west] {\includegraphics[height=\ih]{fig_vpwrist_#1_rgb0.jpg}};
                \node(pl3) at (pl2.east)[anchor=west] {\includegraphics[height=\ih]{fig_vpwrist_#1_mono1.jpg}};
                \node(pr1) at (pl1.south)[anchor=north]{\includegraphics[height=\ih]{fig_vpwrist_#1_mono2.jpg}};
                \node(pr2) at (pr1.east)[anchor=west]  {\includegraphics[height=\ih]{fig_vpwrist_#1_rgb1.jpg}};
                \node(pr3) at (pr2.east)[anchor=west]  {\includegraphics[height=\ih]{fig_vpwrist_#1_mono3.jpg}}; 
                \draw[inner sep=1pt] (p0.north west) rectangle (pr3.south east);
                }
            }
        }
        \pic {aria={124644}{0}{0}};
        \pic {aria={158991}{6.1cm}{0}};
        \pic {aria={189108}{0}{-6.cm}};
        \pic {aria={196609}{6.1cm}{-6.cm}};
    \end{tikzpicture}
    \caption{\textbf{Viewpoints comparison of common body poses}. In each sub-figure, we show the left grayscale camera, RGB camera, the right grayscale camera of Aria glasses (top), left miniAria wristband (middle) and the right miniAria wristband (bottom), respectively.}
    \label{fig:vpcompare}
\end{figure}

\Cref{fig:vpcompare} compares camera viewpoints of Project Aria glasses and miniAria wristbands for 4 common body postures. The sensor suite of Project Aria is detailed in~\cite{aria23surreal}. The miniAria wristbands uses almost identical electronics and sensors, with the exclusion of microphones, barometer, magnetometer and ET cameras. The dynamic range of IMUs on miniAria is increased to count for fast wrist motion. For Project Aria, the left IMU accelerometer saturates at 4g and gyroscope at 500\degree/s, and the right IMU accelerometer saturates at 8g and gyroscope at 1000\degree/s. The saturation range is doubled for both IMUs on miniAria accordingly.

\subsubsection{Recording profile.} 
Project Aria glasses is set to record 30fps RGB video at 1408$\times$1408 pixel resolution, 30fps grayscale videos at 640$\times$480 pixel resolution, 10fps eye tracking videos at 320$\times$240 pixel resolution, 1KHz IMU measurements for the right IMU, 800Hz IMU measurements for the left IMU, 10Hz magnetometer measurements, 50Hz barometer measurements, and 48KHz 7-channel audio. The GNSS, WiFi and BT are turned off for privacy consideration. Similarly, miniAria wristbands record 10fps RGB video at 1408$\times$1408 pixel resolution, 20fps grayscale videos at 640$\times$480 pixel resolution, 1KHz and 800Hz IMU measurements for the right and left IMU respectively. miniAria does not have audio, magnetometer and barometer sensors. The XSens Analyse Pro records 1KHz IMUs and outputs 240Hz full-body motion. 

\section{Data Processing}\label{supp:processing}

\subsection{Project Aria MPS}\label{supp:mps}
Project Aria's machine perception service (MPS) provides building-block algorithms to simplify the processing of the different sensor streams. These functionalities are likely to be available to run on device in real-time for future AR- or smart-glasses. We use the following core functionalities currently offered by Project Aria, and include their raw output as part of the dataset. See~\cite{aria23surreal} and the technical documentation\footnote{\footnotesize{\url{https://facebookresearch.github.io/projectaria_tools/docs/data_formats}}} for more details.

\paragraph{Calibration.} All Aria and miniAria sensors are intrinsically and extrinsically calibrated. For cameras, we use a spherical (equidistant) base model, with additional coefficients for radial, tangential, and thin-prism distortion. Besides this fixed per-device factory calibration, MPS computes time-varying online-calibration as part of the output, that corrects for tiny deformations due to temperature changes or stress applied to the glasses frame.

\paragraph{The 6\,DoF localization.} Every recording is localized precisely and robustly in a common, metric, gravity-aligned coordinate frame, using a state-of-the-art visual inertial odometry (VIO) and simultaneous localization and mapping (SLAM) algorithms. This provides millimeter-accurate 6\,DoF poses for every captured frame and 1\,KHz high-frequency motion in-between camera frames. We provide both a \textit{closed-loop trajectory} that is aligned to this common coordinate frame, as well as the \textit{open-loop trajectory} that is the result of VIO dead-reckoning by strictly causal computing.

\paragraph{Eye gaze.} The gaze direction of the user is estimated as a two outward-facing rays, anchored on the left and right eye respectively. This allows to compute both the direction in which the wearer is looking, as well as -- approximately -- the metric distance at which they are focusing their eyes. We use an optional eye gaze calibration procedure, where the mobile companion app directs the wearer to gaze at a pattern on the phone screen while performing specific head movements. This information was then used to generate a more accurate eye gaze direction, personalized to the particular wearer. 

\paragraph{Point cloud maps.} The 3D point cloud of temporary static scene elements is triangulated from the moving Aria device, using the photometric stereo over consecutive frames and across the left and right monochrome cameras. The output contains the triangulated 3D point clouds, as well as the raw, causally computed, 2D observations of every point in the monochrome camera. The latter allows to compute when each triangulated point is observed -- which can be important to account for, when objects of furniture is moved within scripts. 

\subsection{Motion retargeting}\label{supp:blueman}
\begin{figure}[tb]
    \centering
    \begin{tikzpicture}[inner sep=0pt]
        \def\ih{2.6cm}
        \def\xf{1.4cm}
        \tikzset{
        pics/lingni/.style n args={4}{
            code={
                \node(p0) at(#1,0)[anchor=north west] {\includegraphics[height=\ih]{fig_blueman_#2_xsens.png.transparent.png}};
                \node(p1) at(p0.south)[anchor=north]{\includegraphics[height=\ih]{fig_blueman_#2_xb.png.transparent.png}};
                \node(p2) at(p1.south)[anchor=north]{\includegraphics[height=\ih]{fig_blueman_#2_blueman.png.transparent.png}};
                \node(tt) at(#1, -\ih*3)[anchor=west][text width=\ih, font=\fontsize{6}{10}\selectfont,xshift=#4,yshift=-1ex]{#3};
                }
            }
        } ,
        \pic {lingni={0}{20231023_s0_christopher_green_act1_tsm80q_174881}{F/149/50}{1ex}};
        \pic {lingni={\xf-0.5em}{20230925_s1_xavier_norris_act3_80o04w_087006}{F/155/64}{1em}};
        \pic {lingni={\xf*2-1em}{20231113_s1_greg_clark_act4_chhp0x_160253}{M/164/88}{3ex}}; 
        \pic {lingni={\xf*3-0.4em}{20230726_s1_thomas_nixon_act2_bsfnuw_083687}{F/168/59}{2ex}}; 
        \pic {lingni={\xf*4-1em}{20230803_s0_robert_howard_act2_kobpfk_117766}{F/173/56}{3ex}};
        \pic {lingni={\xf*5-0.5em}{20231108_s1_kurt_young_act4_t5o4jk_106430}{M/175/73}{4ex}};
        \pic {lingni={\xf*6+1ex}{20230925_s0_suzanne_romero_act4_82fp8g_082273}{M/188/83}{4ex}};
        \pic {lingni={\xf*7+2ex}{20230721_s0_mitchell_mcdonald_act4_mq1a2s_025027}{M/199/69}{4ex}};
    \end{tikzpicture}
    \caption{\textbf{XSens retargetting for different body profiles.} Each column shows the fitting results of different body shape profile. This is specified by the bottom text, where the tuple X/Y/Z specifies a X-gender subject with Ycm height and ZKg weight. The top row shows the XSens skeleton with 79 anatomical landmarks as defined in~\cite{mvnmanual}. The landmarks are colored-coded by the fitting errors, with the darker color the lower error. The second row shows the retargetted mesh in polygons. The last row shows the same mesh rendered with normal shading.}
    \label{fig:blueman-fit}
\end{figure}
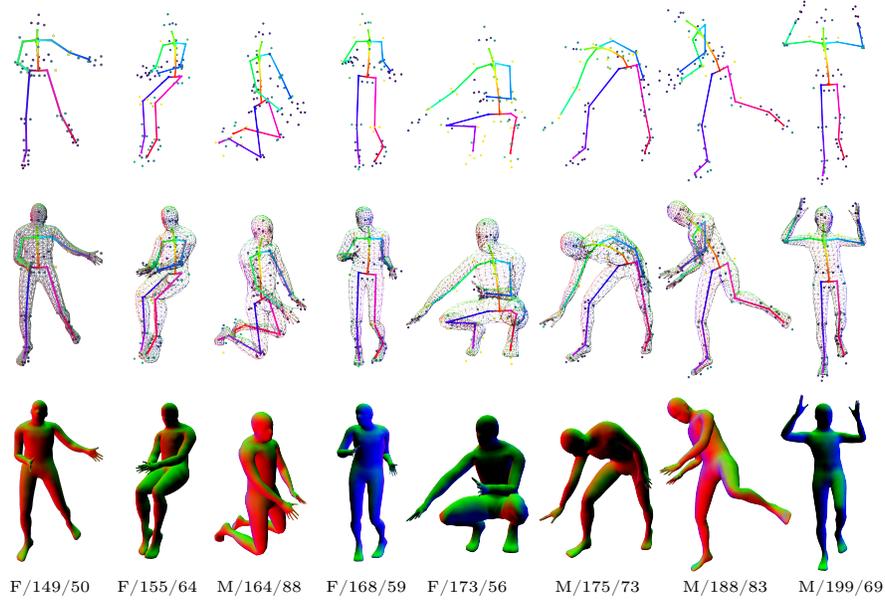

Here we present details on the human model used to represent the ground truth body motion, and the optimization solver for motion retargeting from XSens output. Examples of retargeting results are shown in \cref{fig:blueman-fit}. We will release the model and the solver library upon dataset release. 

\begin{figure}[tb]
    \centering
    \begin{subfigure}{0.35\linewidth}
    \includegraphics[width=\linewidth]{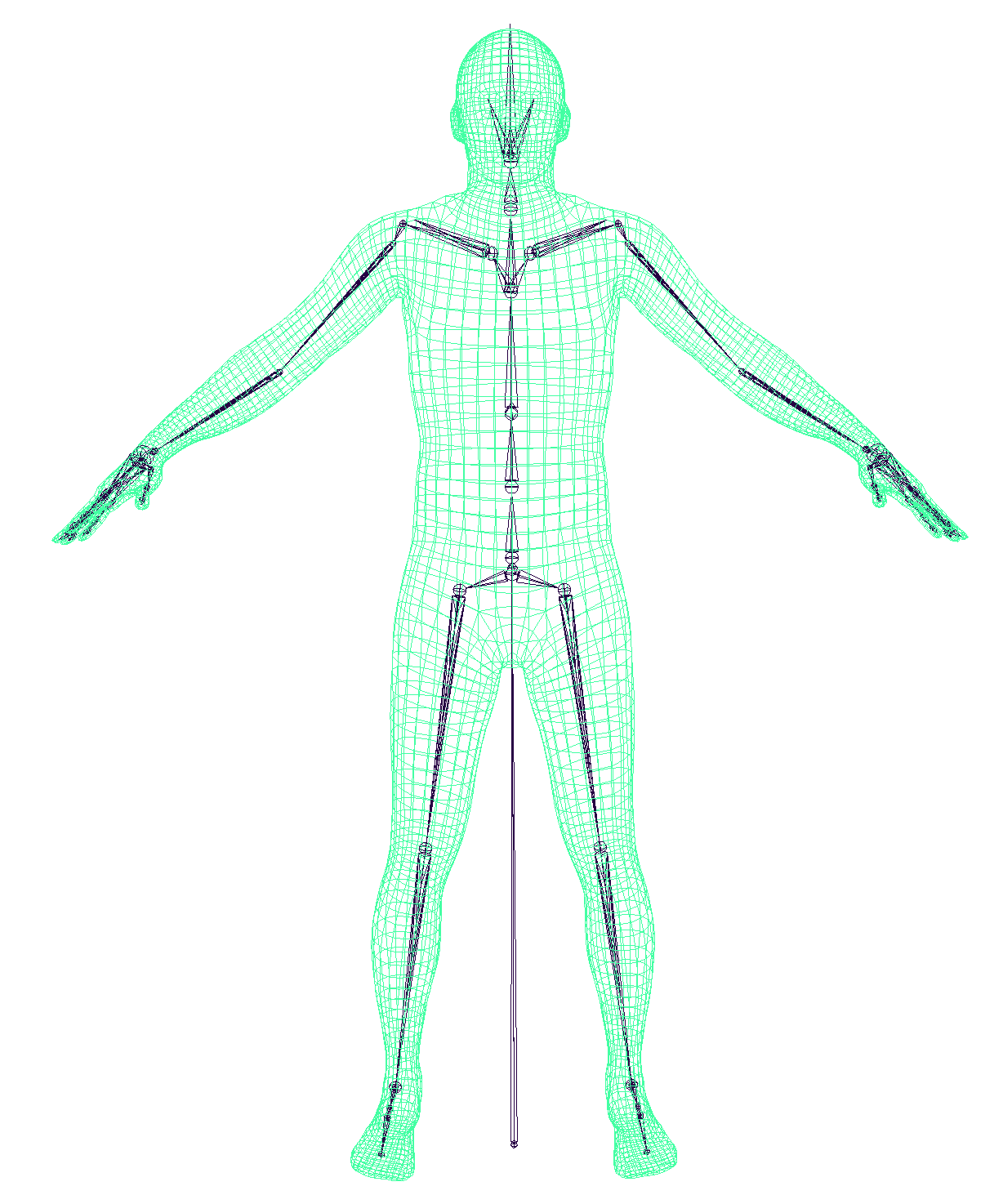}
    \end{subfigure}
    \hfill
    \begin{subfigure}{0.35\linewidth}
    \includegraphics[width=\linewidth]{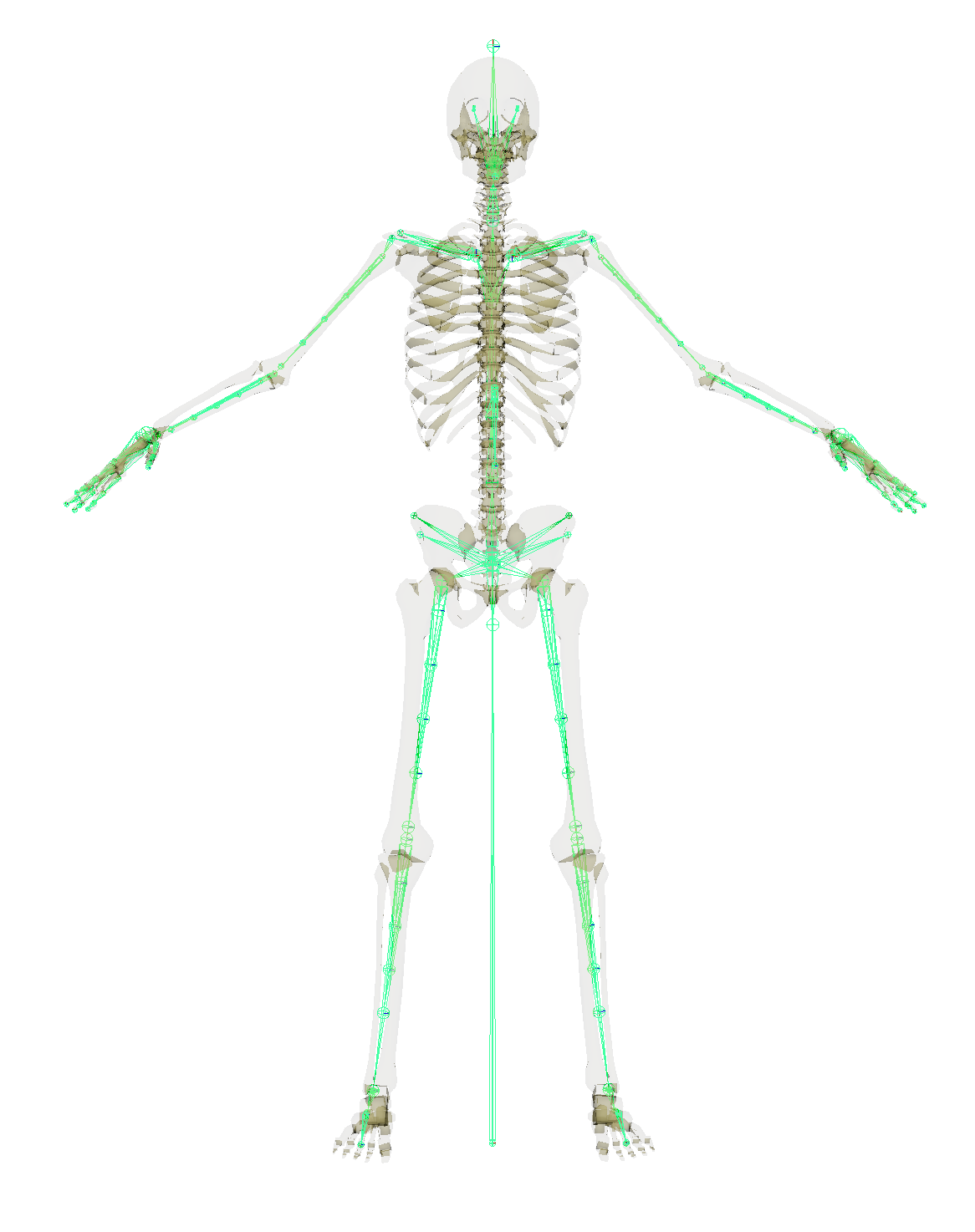}
    \end{subfigure}
    \hfill
    \begin{subfigure}{0.213\linewidth}
    \includegraphics[width=\linewidth]{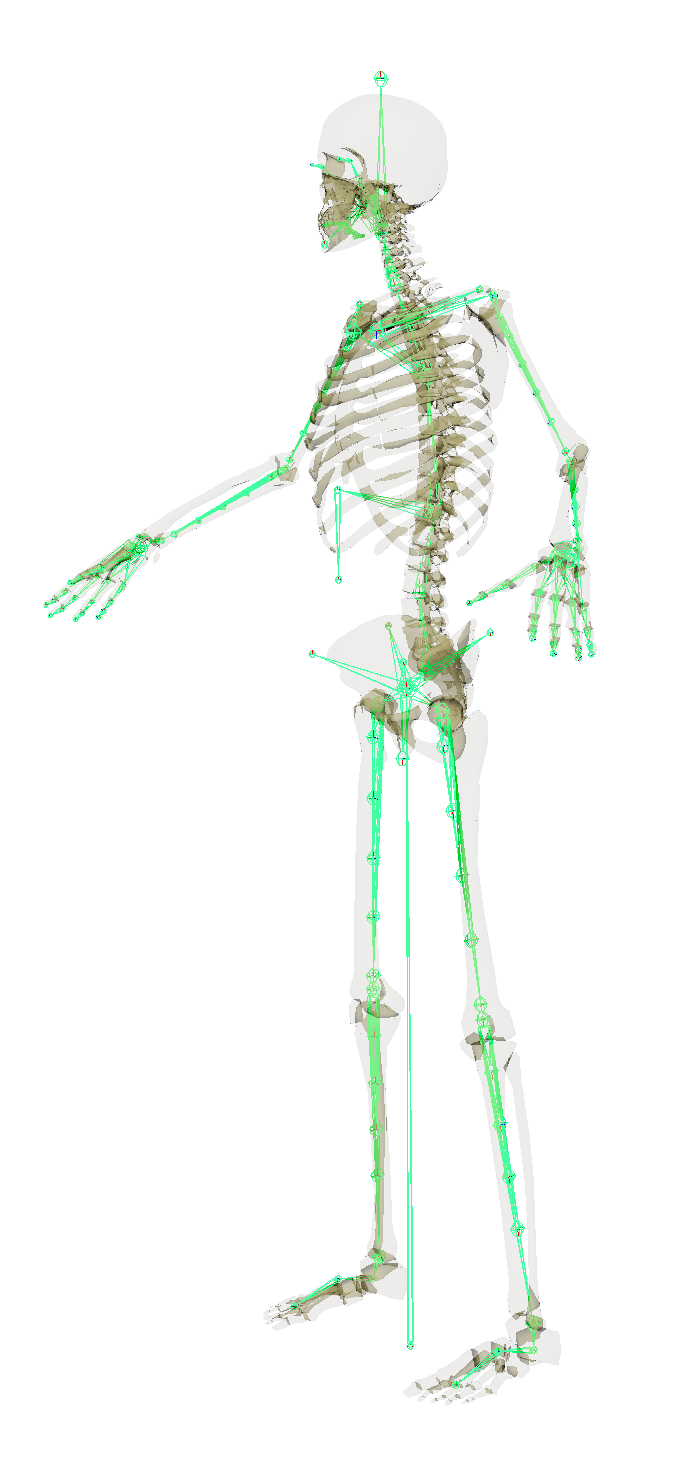}
    \end{subfigure}
    \hfill
    \caption{\textbf{Our anatomically inspired human model.} Its joints are designed and placed to follow human anatomy. Left: frontal view of the skeleton (dark) and the body quadmesh (green). Middle: frontal view of the skeleton overlay on a skeleton model. Right: side view of the skeleton model.}
    \label{fig:body-model}
\end{figure}

\paragraph{Human model.} Our human model showing in \cref{fig:body-model} (left) consists of 159 skeletal joints that deform a manifold quadrangle mesh through linear blend skinning (LBS). Among them, 28 joints control the face, such as eyes, jaw, and the tongue; and 42 joints control all fingers on both hands, 21 joints on each side. The remaining 89 joints that control the body are what we use in retargeting to represent the body motion ground truth, while leaving the face and finger joints unchanged. 

Among the 89 body joints, we use one joint to represent the global transformation in the world. 35 joints are designed to represent bones in the human body and they loosely follow the human anatomical structure (\eg the forearm joint corresponds to the ulna and serves as the twist pivot of the wrist, which is located at the tip of what would be the radius). \Cref{fig:body-model} (middle, right) overlays the joints on a skeleton model for reference. In addition, we designed 53 "helper" joints for the purpose of improving skinning deformation, as commonly done in computer graphics. These helper joints are driven by the 35 bones with a simple linear relationship. For example, there are five helper joints along each upper and lower limb respectively. Their purpose is to distribute the twist rotation values between the two end joints along the limbs, to avoid the "candy wrapper" artifact of linear blend skinning.

\paragraph{Model parameterization.} The body model is parameterized by identity parameters $\boldsymbol{\phi} \in \mathbf{R}^{12}$ and pose parameters $\boldsymbol{\theta} \in \mathbf{R}^{58}$. They are designed manually from heuristics to follow anatomical principles while reducing mode complexity. The identity parameters consist of a global scale of the model, three end effector scales (\ie hand, hands, and feet), and eight translations that represent bone lengths symmetrically for both sides (\ie hip width, spine length, neck length, shoulder width, upper arm length, lower arm length, upper leg length, lower leg length). The translation values offset the position of a joint in its local space (\ie parent space). The 58 pose parameters consist of 6\,DoF for the global transformation, and 52 euler angles that represent local joint rotations. This is a reduced set of rotations for the 35 joints since we model limits and synergies of joints. For example, the elbows and knees only have one rotational DoF. And we control the four spine joints with only two groups of 3D rotations, which co-activate adjacent spine joints with a fall off for better postures. The specific definitions of these parameters will be released with the model. 

Since XSens provides only skeletal poses with no body shape information, we do not change the body mesh from the template for an individual, except for skinning deformations from the identity parameters.

\paragraph{Optimization solver.} As described in the paper, we solve an inverse kinematics optimization problem using anatomical landmarks from XSens. We manually define the location of these landmarks on our human model and parent them rigidly under appropriate joints. We don't expect our manual definition to be accurate, so we only use them as an initial guess, and aim to solve the local landmark locations in the optimization. For every motion, XSens outputs the global locations of these landmarks, which we take as input,
and solve for the identity parameters of the user, the landmark offsets, and the per-frame pose parameters using Equation (1). This non-linear least-square optimization is solved using the Levenberg–Marquardt algorithm.

The data from XSens often contain various artifacts. Two main sources of error are inaccurate body dimension measurements, and the simplified skeleton model XSens uses to solve for body motions from their sensor measurements. They result in body self-penetrations, especially between the hands and the body, and unnatural poses. To mitigate these artifacts, we incorporate two additional loss terms as the optimization objective: parameter limits and body collisions penalty. Parameter limits are defined as quadratic falloffs at the manually designed boundaries. Body collisions are computed from manually defined collision proxies (\ie tapered capsules) for each body part. At every optimization step, we compute pairs of colliding bodies, and penalize the penetration distance. Optionally, we can also use a smoothness objective to penalize change of pose parameters between consecutive frames. Because these objectives are also expressed as least-square functions, they can be optimized using the Levenberg–Marquardt algorithm as well.

In practice, a motion sequence can be several minutes long and it is difficult and expensive to solve the entire sequence all together. We instead solve Equation (1) from only a subset of uniformly sampled frames (\eg 200 frames) from a long sequence to obtain the identify parameters and landmark offsets. We then use them to solve for pose parameters for the entire sequence frame-by-frame, which is much simpler and faster. 

\subsection{In-context motion narration}\label{supp:narration}

\begin{figure}[tb]
    \centering
    \includegraphics[width=\textwidth]{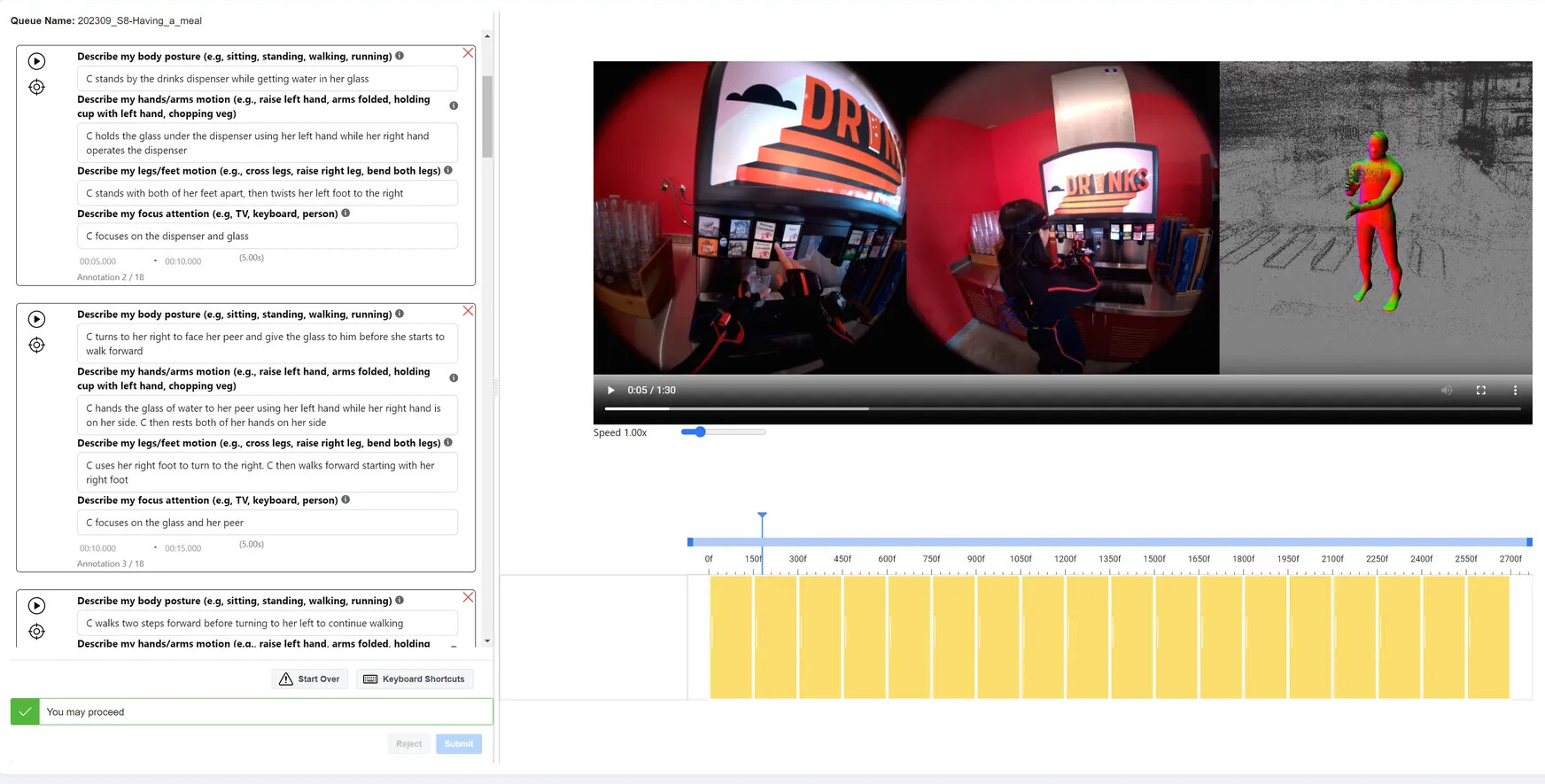}
    \caption{\textbf{The narration tool interface.} Top right: annotators are presented with the synchronized egocentric RGB video, third-person RGB video and the rendering of the full-body motion. Bottom right: the video segmentation bar, where each yellow bar is one clip created by annotators to add description. Left: text boxes to enter answers to predefined questions. There is one text box per created segment. Following the Ego4D convention~\cite{ego4d22cvpr}, participants are referred by C. }
    \label{fig:narration-ui}
\end{figure}

We employ 25 annotators to write text descriptions in English for the dataset. Compare to the existing motion-language narration, our goal is to obtain in-context annotations that aligns full-body motion, egocentric perception, and language by offering descriptions about body poses and how the wearers interact with objects, environments and other people. 

\subsubsection{Tool interface.} 
To ground the narration, we develop the annotation tool to show annotators the synchronized videos of the egocentric RGB camera, the third-person RGB camera and the full-body motion rendered with scene point clouds. The tool interface is shown in \cref{fig:narration-ui}. The tool allows annotators to segment the video into arbitrary length of clips. Then for each video segment, annotators write full-sentence answers to a set of predefined questions. We identify three annotation tasks to control the granularity from coarse to fine. The same tool interface is used to produce annotation for each tasks, where the questions are altered.

\subsubsection{Guidelines and narration questions.} 
We train annotators to identify the participants from the appearance of mocap suit and 3D rendering. Annotators are required to give clear descriptions as if they are describing the scenario over the phone for the recipient to reconstruct the scene. Annotators are asked to make reasonable segments that aligns with motion transition, and their descriptions should include all actions during that segment. 

\paragraph{Motion narration} is the finest level of annotation, which focuses on detailed body poses. For this task, each segment is between 3-5s long where annotators answer the following questions about the participants, \ie 1) ``describe the full-body motion including the direction/intent of the motion and environment''; 2) ``describe the hands/arms motion''; 3) ``describe the detailed legs/feet motion'' and 4) ``describe the focus attention''. 
\paragraph{Atomic actions} is the next level. For this task, segments are between 3-5s duration as well, however annotators only answer 1 question, \ie ``describe the actions, including body posture, direction/intend of motion, and interaction with objects and other people''. Compare to the previous task, annotators are encouraged to use verbs to describe actions whenever possible instead of focusing on body poses.
\paragraph{Activity summarization} is the coarsest level of annotation, where annotators create segments of 15-30s duration and answer 1 question, \ie ``summarize the major activity, including the interaction with people/objects and the scene''.

\begin{table}[h]
    \centering
    \small  
    \setlength{\tabcolsep}{1.5pt}
    \newcolumntype{C}[1]{>{\centering\arraybackslash}p{#1}}
    \newcolumntype{R}[1]{>{\raggedleft\arraybackslash}p{#1}}
    \newcolumntype{L}[1]{>{\raggedright\arraybackslash}p{#1}}
    \begin{tabular}{L{3.5cm} *{3}{R{7mm}} *{6}{R{1.1cm}} }
    \toprule
    script                  & i.d.        &o.d.        & 2pt.       &pct/\%    &dur./h  &H/Km      &L/Km       &R/Km   \\ \toprule
    S1 Relax at home        & \checkmark  &            &            & 2.40     &9.02    &  2.97    &  3.11     &  3.59 \\ \hline
    S2 Where is X           & \checkmark  &            & \checkmark & 9.80     &36.87   & 30.25    & 37.90     & 41.49 \\ \hline
    S3 Welcome to my place  & \checkmark  & \checkmark & \checkmark & 5.00     &18.80   & 57.94    & 64.89     & 66.44 \\ \hline
    S4 Body stretch         & \checkmark  &            &            & 2.00     &7.62    & 22.80    & 27.93     & 26.92 \\ \hline
    S5 Cardio/workout       & \checkmark  &            &            & 2.20     &8.39    &  5.03    &  8.42     &  8.68 \\ \hline
    S6 Dancing              & \checkmark  &            &            & 1.70     &6.55    &  9.62    & 28.32     & 26.29 \\ \hline
    S7 Cooking              & \checkmark  &            &\checkmark  & 12.90    &48.52   &  7.46    & 14.45     & 13.31 \\ \hline
    S8 Having a meal        & \checkmark  & \checkmark &            & 4.45     &16.73   & 33.35    & 41.03     & 47.47 \\ \hline
    S9 Making a mess        & \checkmark  &            &            & 1.83     &6.90    & 14.09    & 15.18     & 15.21 \\ \hline
    S10 Housekeeping        & \checkmark  &            &\checkmark  & 6.80     &25.57   &  8.98    & 12.19     & 13.24 \\ \hline
    S11 laundry and packing & \checkmark  &            &            & 2.41     &9.05    & 10.26    & 14.46     & 14.75 \\ \hline
    S12 Game night          & \checkmark  &            &\checkmark  & 10.39    &39.08   & 21.60    & 22.90     & 29.58 \\ \hline
    S13 Charades            & \checkmark  &            &            & 2.43     &9.15    &  7.47    & 14.63     & 15.27 \\ \hline
    S14 By the desk         & \checkmark  &            &            & 1.52     &5.73    &  2.39    &  2.25     &  2.46 \\ \hline
    S15 Do as I command     & \checkmark  &            &            & 2.06     &7.74    &  8.77    & 15.60     & 15.31 \\ \hline
    S16 Simon says          & \checkmark  & \checkmark &            & 15.72    &59.15   & 61.21    & 73.73     & 78.85 \\ \hline
    S17 In the office       & \checkmark  &            &            & 1.90     &7.16    &  5.31    &  6.80     &  7.92 \\ \hline
    S18 Hike                &             & \checkmark & \checkmark & 2.77     &10.44   & 27.33    & 30.68     & 30.50 \\ \hline
    S19 Fresh air           &             & \checkmark & \checkmark & 5.76     &21.66   & 56.17    & 67.98     & 72.77 \\ \hline
    S20 Party time          & \checkmark  &            & \checkmark & 5.86     &22.04   & 17.87    & 21.84     & 24.20 \\ \midrule      
    sum                     &18           &5           &8           &100.0     &376.2   & 411.0    &524.4      & 554.4 \\   
    \bottomrule
    \end{tabular}
    \caption{\textbf{Summary of scenarios.} We mark scenarios for capturing indoor activities (i.d.), outdoor activities (o.d.), and two-participants collaborations (2pt). The recording hours and trajectory length is broke down by head(H), left wrist(L) and right wrist(R). The percentage of each scenario is computed from the overlapping recording time of all devices. The recording duration in this table also include time spent on eye calibrations at the start of each recording and additional data without low-quality motion. This is longer than \hourcnt hours, which is the amount of clean data with all modalities.}
    \label{tab:scenario}
\end{table}

\section{Dataset Details}\label{supp:facts}
\subsection{Recording scenarios}\label{supp:scenario}
\subsubsection{Definitions.}
The \nymeria dataset defines \scriptcnt scenarios of daily indoor and outdoor motion. They are defined as follows.
\begin{itemize}
    \item \textit{S1 Relaxing at home.} 
    In this script participants pretend to have a relaxing evening at home after work. The common activities include sitting on couch or laying on sofa to watch TV, finding books to read, looking for snacks or beverage to enjoy etc. Operators are instructed to prompt participants to change their activities if participants have been staying at one location for more than 5min. 
    \item \textit{S2 Where is X.}
    In this script, participants look for objects left behind in the house. The common activities include walking around the house, standing on toes or bending over to open cabinets and drawers, looking around, communicating with operators etc. Participants usually found 5 - 10 objects during a 15min recording. Operators are instructed to hide objects at places where people commonly forgot them behind, \eg phone under pillow, laptop next to nightstand, keys in jacket pockets, toys under the bed etc. 
    \item \textit{S3 Welcome to my home.}
    In this script, participants pretend to be the homeowners to give visitors a house tour. The common activities include greeting guests, offering beverages, walking around the house, having natural conversations etc. When there are two participants, they take turns to introduce different rooms. Operators and observers are instructed to prompt participants by asking for water, to see different rooms and questions about the house decorations etc. 

    \item \textit{S4 Body stretch. }
    In this script, participants follow video instruction to perform light body stretch exercises. We rotate different youtube videos to add some diversity. A common stretching exercise is yoga alike motion. We always confirm with participants for their comfortable level of exercising before proceeding.
        
    \item \textit{S5 Cardio/workout.} Similar to S4, in this script participants follow video instruction to for workout sessions. Compare to S4 which is designed to record full body stretching, S5 captures faster body motions, \eg jumping jacks, kicking feet back into a plank position, squatting etc. A few houses are equipped with indoor biking and treadmill, where participants are encouraged to leverage these workout tools instead of following videos. We always confirm with participants for their comfortable level of exercising prior to recording.
    
    \item \textit{S6 Dancing.} Similar to S4 and S5, in this script participants mainly follow video instructions to dance. To capture the natural reactions, participants are not required to know how to dance prior to recording. we always confirm with participants for their comfortable level before proceeding. Compare to S4 and S5, S6 captures more frequent arm swings and body rotation. We choose multiple dancing videos to include salsa, cha cha cha, rumba etc.
    
    \item \textit{S7 Cooking.} In this script, participants are provided with cooking ingredients to prepare a dish in the kitchen. The common activities include gathering ingredients from fridge, looking for tools, washing and chopping vegetables, stirring or frying with pans, seasoning, cleaning afterwards etc. The commonly used receipes for cooking include quesadilla, tacos, stir fry, baking, salad, BLT sandiches etc.
    
    \item \textit{S8 Having a meal.} In this script, participants have lunch with operators. We capture this scenario at houses and in an large open-space cafeteria with outdoor patio. The common activities include setting table, getting food and drinks from buffet, having meal with friends, returning dishes, cleaning up table, load/unload dishwasher etc. 
    
    \item \textit{S9 Making a mess.} This script is defined for participants to create natural messy home. Typical activities include walking around the house, carrying objects and misplacing them, throwing stuff around etc. 
    
    \item \textit{S10 Housekeeping.} This script is usually done in combination with S9, where participants clean up the house. It is usually done by a different participant who created messy home before. The common activities include ordering objects, making bed, cleaning up floor etc. 

    \item \textit{S11 Laundry and packing.} In this script, participants are asked to do laundry and packing for travel. The common activities include folding clothes, load/unload washing machine, hanging up clothes, packing suitcase etc. 
    
    \item \textit{S12 Game night.} In this script, participants playing various games with other people. We also make use of existing entertaining facilities whenever possible. The common activities captured include playing poker, chess, gengar, boardgames, puzzels as well as fussball, pacman, pooling, mini golf, dart etc.
    
    \item \textit{S13 Charades.} Similar to S12, this script is also defined for participants to play games with other people, however, the focus is to act with body language with participants being the performers. Depending on the characters of participants, either participants come up with various themes to act themselves for the operators to guess or participants are prompted by the operators. Operators are instructed to encourage participants to exaggerate their body language and move around.   
    
    \item \textit{S14 By the desk.} In this script, we focus on how people doing work from a working from home setup. To simulate the real-life scene, participants are often prompted to do computer-related tasks, \eg typing speed test, solving online quiz, browsing websites etc. We also capture participants writing or doodling on a notebook and doing crafting such as origami.  
    
    \item \textit{S15 Do as I command.} This is the only script designed to cover a set of useful locomotion for algorithms to derive full-body pose prior instead of to understand natural daily activities. Participants are asked to act according to a predefined motion list, including walking, jogging, running, skipping, jumping etc on the spot, in a circle and backwards, rotating head/upper body/arms/ankle clockwise and anti-clockwise, kicking legs, bowling, touching toes, boxing, squatting, sitting down, taking stairs, lying flat etc.
    
    \item \textit{S16 Simon says.} This script shares the same spirit of S15, where participants follow the instructions from operators to perform actions. However, the goal is tailored towards a mini real-life events, \eg making tea, taking picture off the wall, measuring furniture, water plants, bring grocery from the car, picking flowers from the garden etc. As the name suggest, the scenario is inspired by the children's game, where a person always uses the phrase ``Simon says'' to propose a action for everyone else to follow. This script helps us better capture the long-tailed daily activities that are otherwise difficult to incorporate into other recording scenarios. This is also one of the scenarios where audio records the psuedo ground truth of action labels. 
    
    \item \textit{S17 In the office.} This script is captured in an office building with multiple meeting rooms. Compare to S14, this scenario focuses on working onsite. The common activities include navigating in the building, taking stairs, having conversation at different locations, finding meeting spaces, giving presentations, using whiteboard, working on a laptop, taking breaks etc.
    
    \item \textit{S18 Hike.} In this script, participants hike in the woods. We include multiple hiking trails with easy flat ones and median hilly ones.  
    
    \item \textit{S19 Fresh air.} This script focuses on outdoor refreshing activities, where participants typically play badminton, soccer, swing, jogging, or biking. We record most scenarios on a campus and in the backyard of houses.
    \item \textit{S20 Party time.} This script is design for people to decorate a house for party or holiday celebrations. The common activities include making balloon (animals), setting up table, hanging decorations, arranging furniture etc. Most of our party scenario are recorded with two participants.
\end{itemize}

\subsubsection{Statistics.}
The summary of recording statistics is given in Table~\ref{tab:scenario}. Note that the recording duration in \cref{tab:scenario} includes time spent on in-session eye calibration at the beginning of each recording and additional data where motion recording is poor quality. Therefore the summed value is longer than \hourcnt hours, which is the amount of clean data with all modalities aligned with high-quality motion.

\subsection{Locations}\label{supp:locations}
\begin{figure}
    \centering
    {
        \setlength{\fboxsep}{0pt}\setlength{\fboxrule}{0.5pt}
        \begin{minipage}{0.517\linewidth}%
        \fbox{\includegraphics[width=\linewidth]{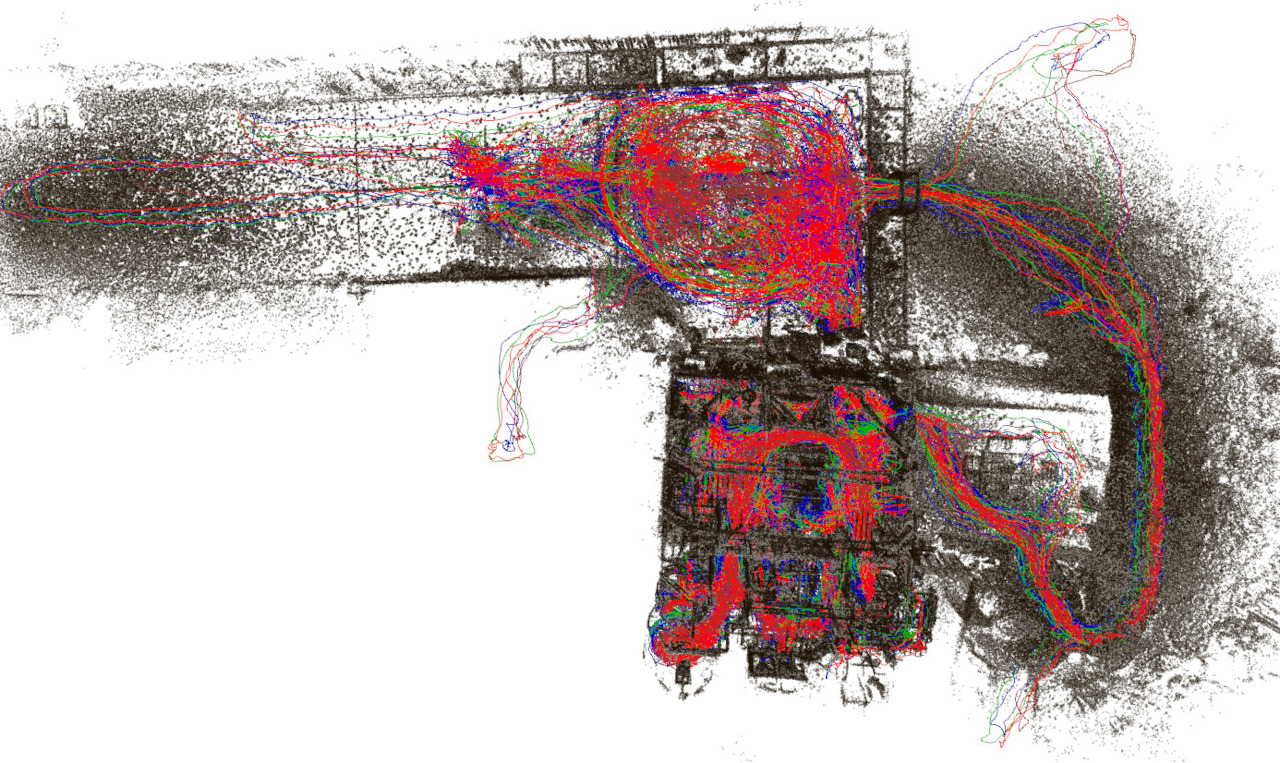}}%
        \end{minipage}~%
        \begin{minipage}{0.15\linewidth}%
        \fbox{\includegraphics[width=\linewidth]{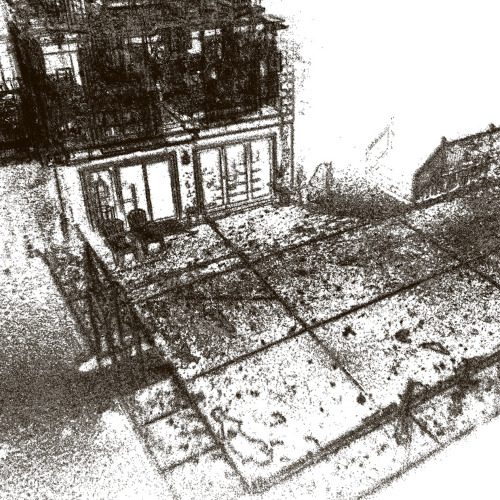}}\\[0.05mm]%
        \fbox{\includegraphics[width=\linewidth]{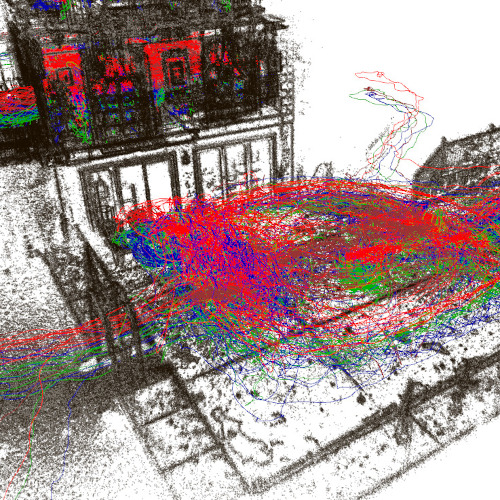}}%
        \end{minipage}~%
        \begin{minipage}{0.15\linewidth}%
        \fbox{\includegraphics[width=\linewidth]{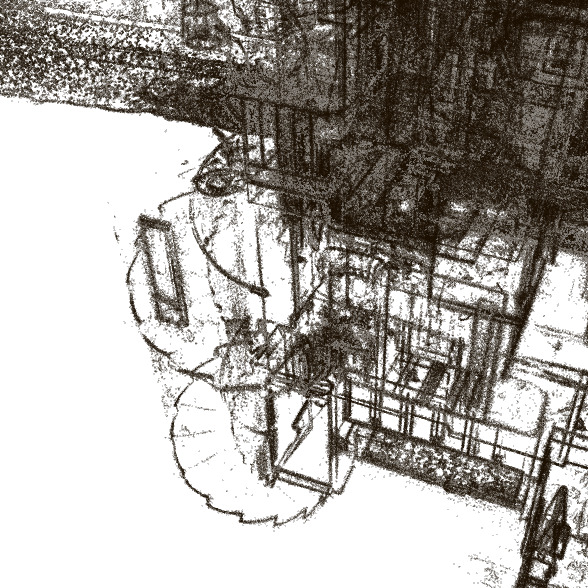}}\\[0.05mm]%
        \fbox{\includegraphics[width=\linewidth]{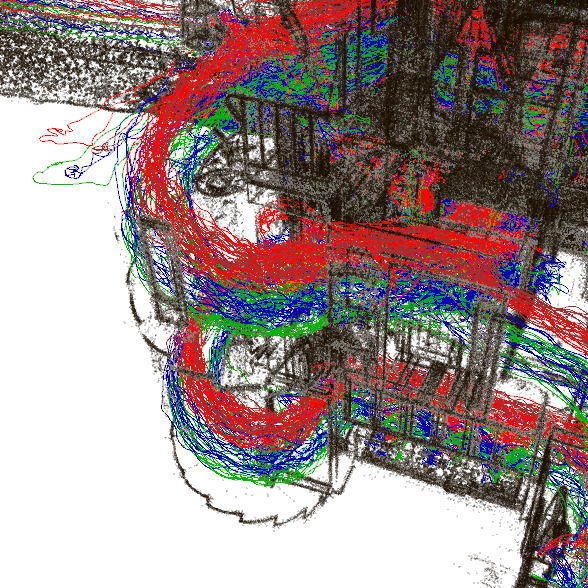}}%
        \end{minipage}~%
        \begin{minipage}{0.15\linewidth}%
        \fbox{\includegraphics[width=\linewidth]{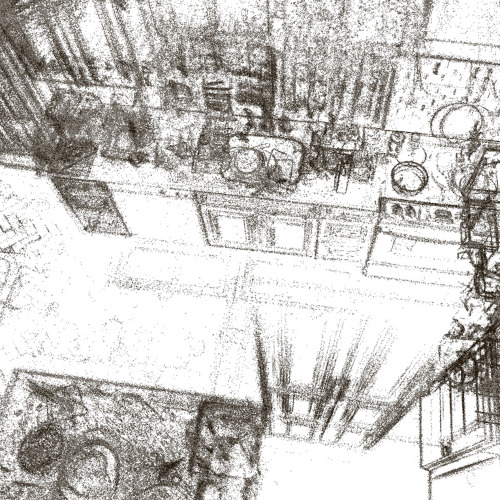}}\\[0.05mm]%
        \fbox{\includegraphics[width=\linewidth]{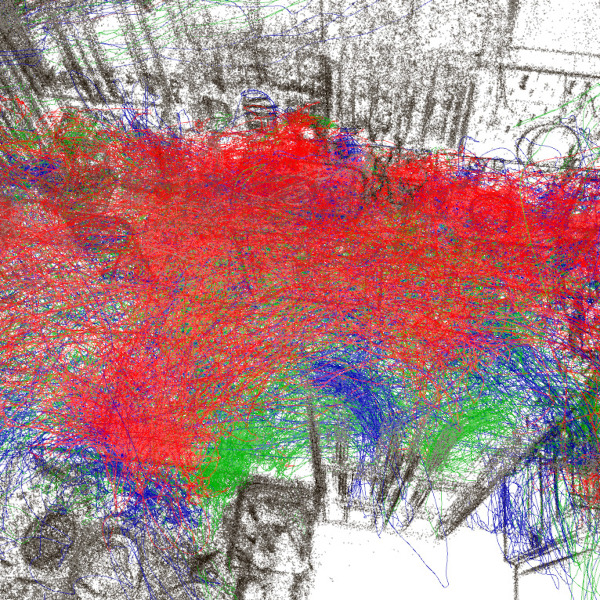}}%
        \end{minipage}~%
    }\\
    {
        \setlength{\fboxsep}{0pt}\setlength{\fboxrule}{0.5pt}
        \begin{minipage}{0.517\linewidth}%
        \fbox{\includegraphics[width=\linewidth]{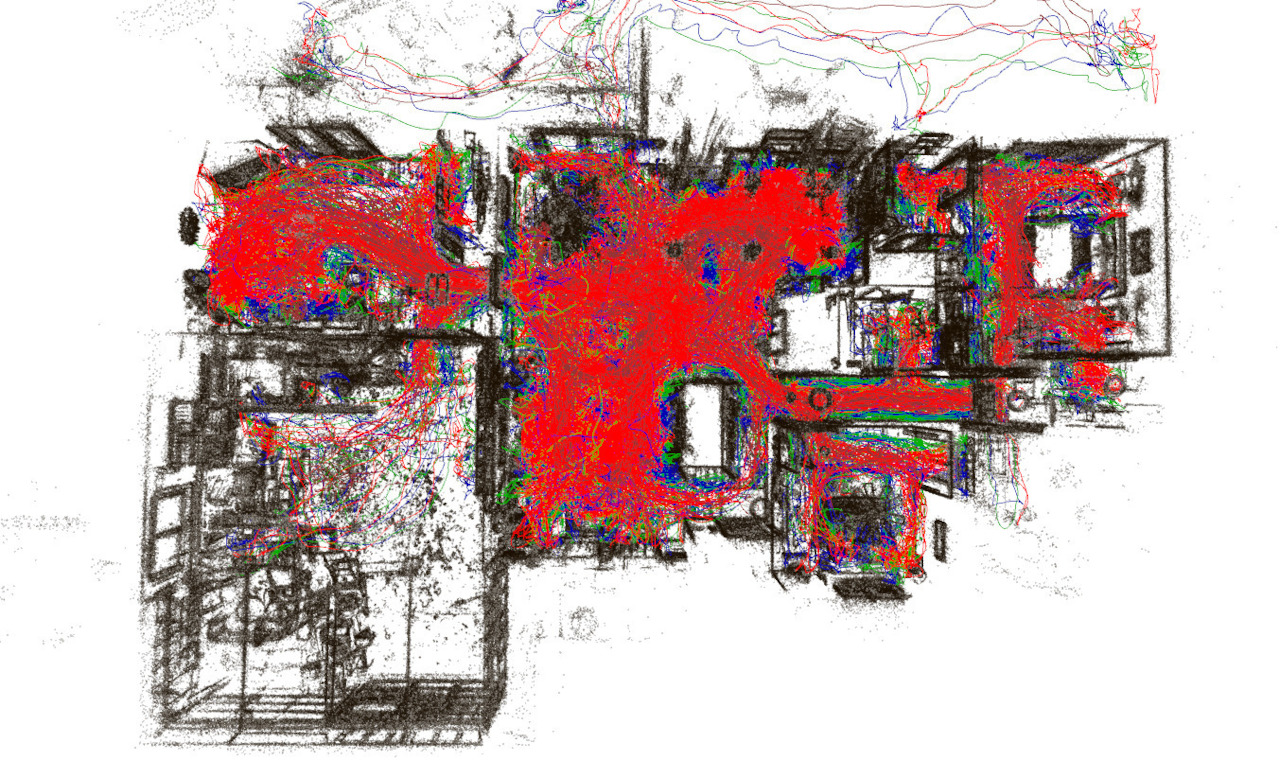}}%
        \end{minipage}~%
        \begin{minipage}{0.15\linewidth}%
        \fbox{\includegraphics[width=\linewidth]{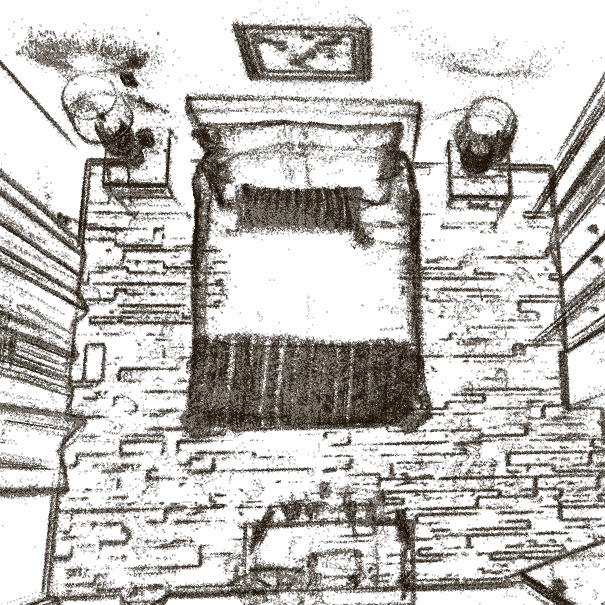}}\\[0.05mm]%
        \fbox{\includegraphics[width=\linewidth]{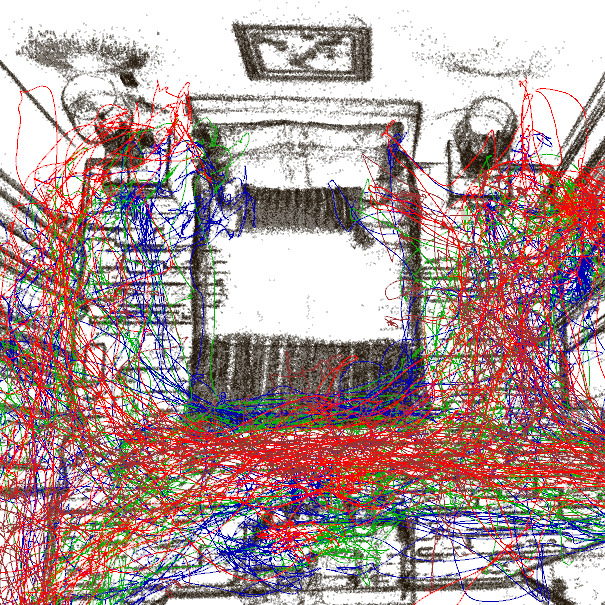}}%
        \end{minipage}~%
        \begin{minipage}{0.15\linewidth}%
        \fbox{\includegraphics[width=\linewidth]{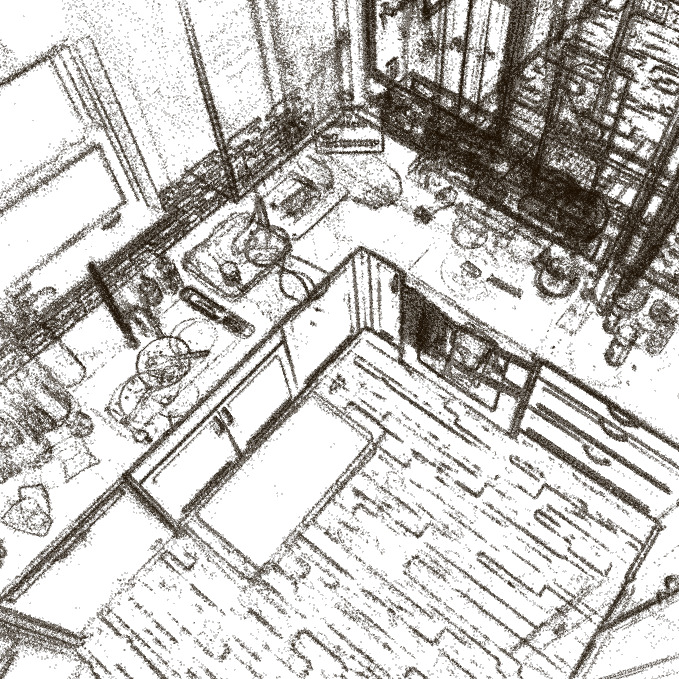}}\\[0.05mm]%
        \fbox{\includegraphics[width=\linewidth]{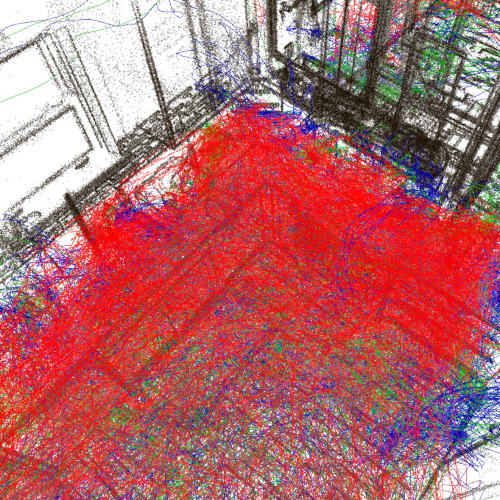}}%
        \end{minipage}~%
        \begin{minipage}{0.15\linewidth}%
        \fbox{\includegraphics[width=\linewidth]{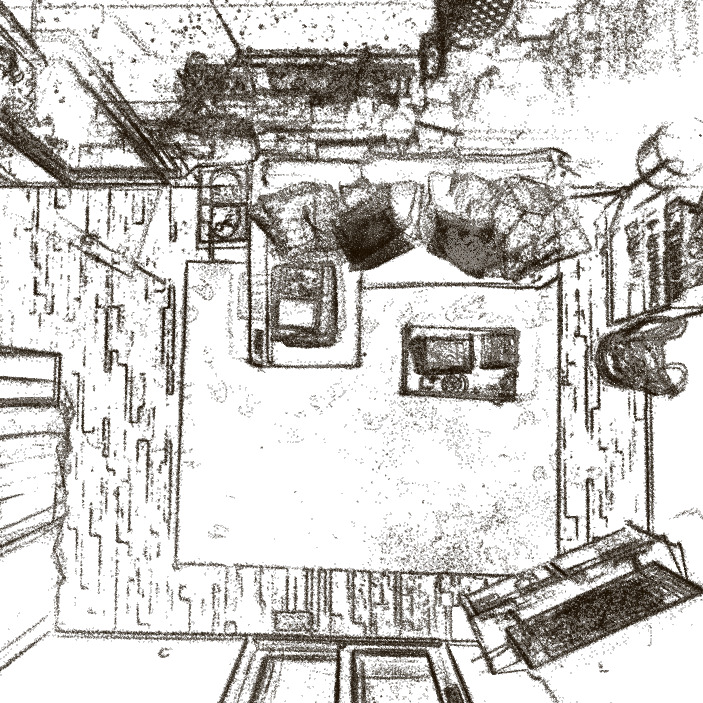}}\\[0.05mm]%
        \fbox{\includegraphics[width=\linewidth]{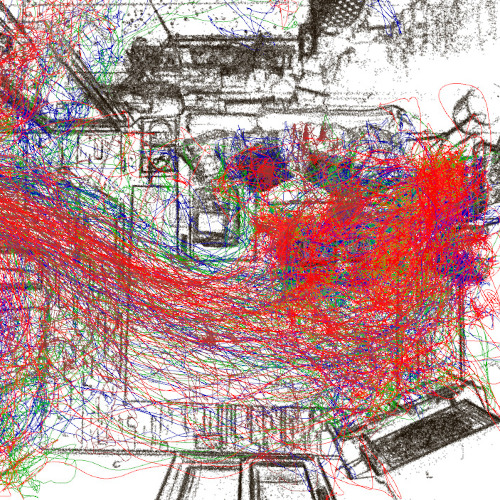}}%
        \end{minipage}~%
    }\\
    {
        \setlength{\fboxsep}{0pt}\setlength{\fboxrule}{0.5pt}
        \begin{minipage}{0.517\linewidth}%
        \fbox{\includegraphics[width=\linewidth]{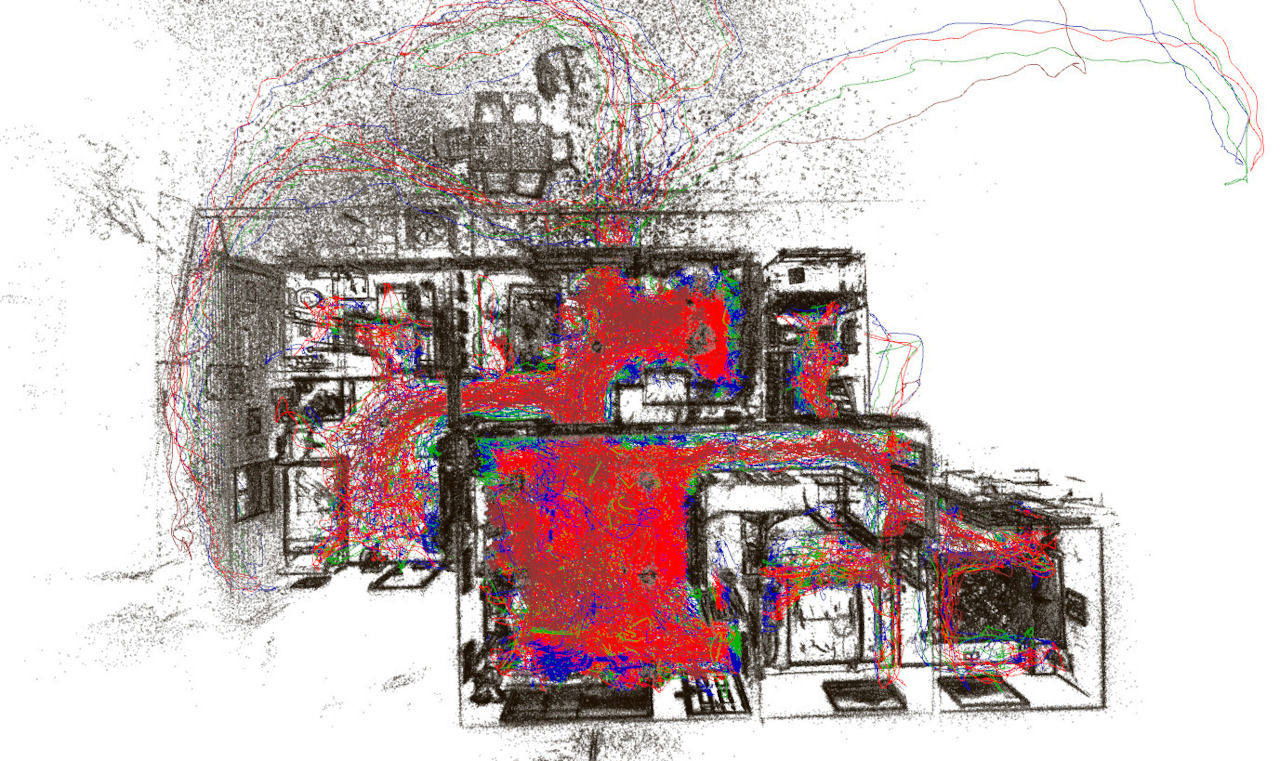}}%
        \end{minipage}~%
        \begin{minipage}{0.15\linewidth}%
        \fbox{\includegraphics[width=\linewidth]{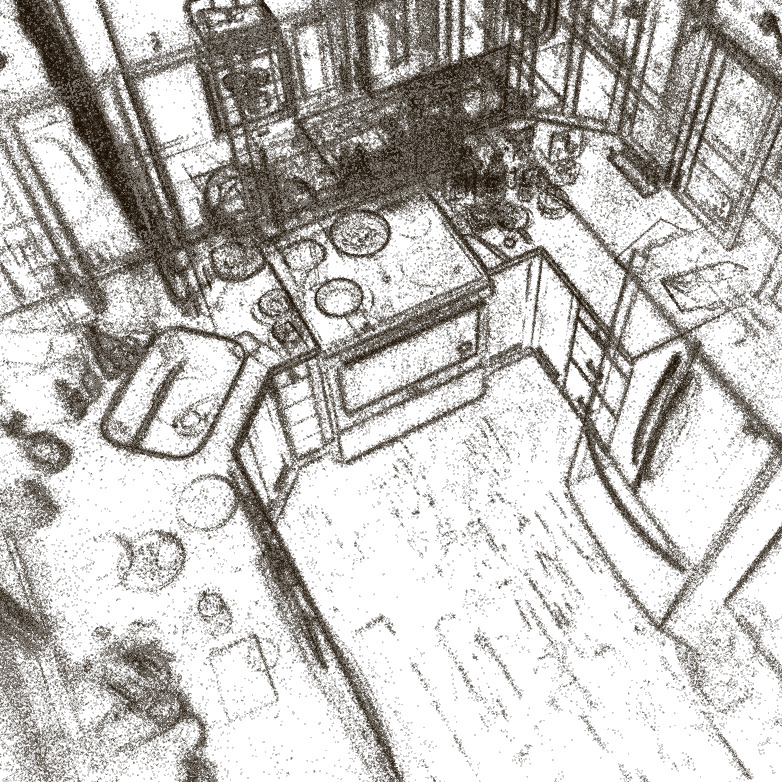}}\\[0.05mm]%
        \fbox{\includegraphics[width=\linewidth]{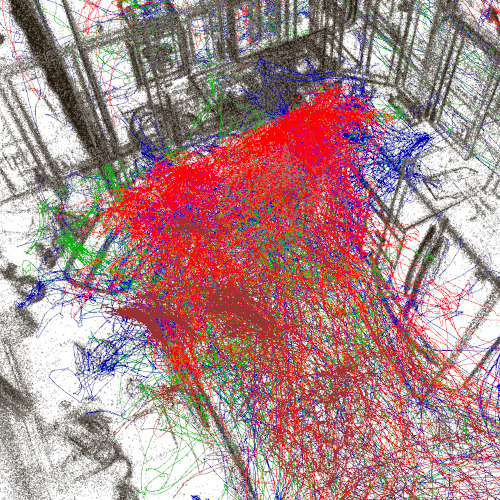}}%
        \end{minipage}~%
        \begin{minipage}{0.15\linewidth}%
        \fbox{\includegraphics[width=\linewidth]{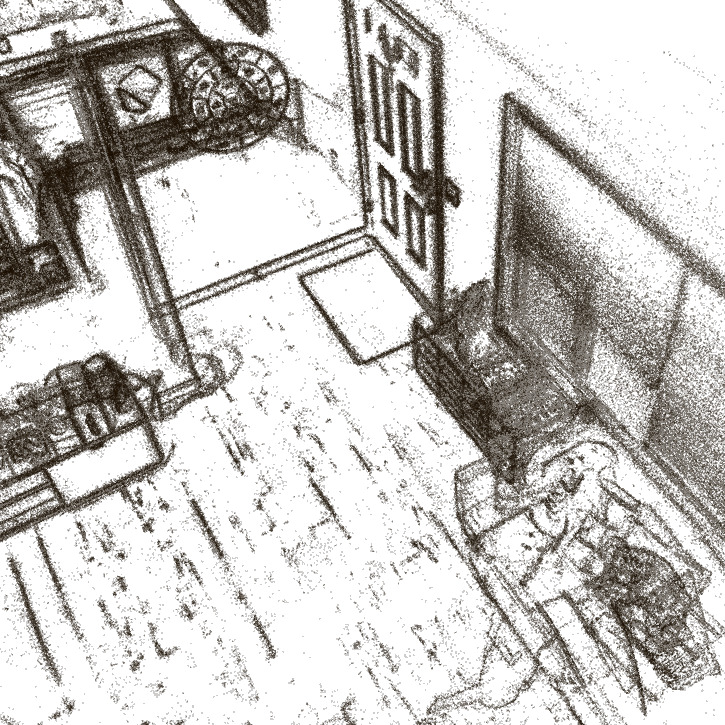}}\\[0.05mm]%
        \fbox{\includegraphics[width=\linewidth]{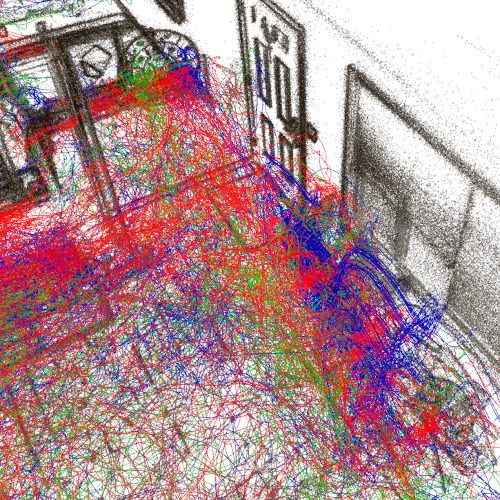}}%
        \end{minipage}~%
        \begin{minipage}{0.15\linewidth}%
        \fbox{\includegraphics[width=\linewidth]{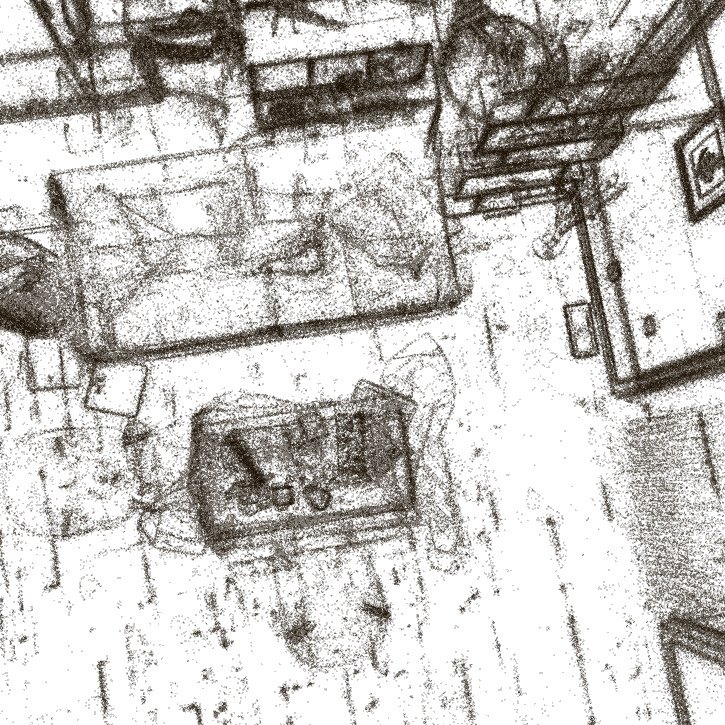}}\\[0.05mm]%
        \fbox{\includegraphics[width=\linewidth]{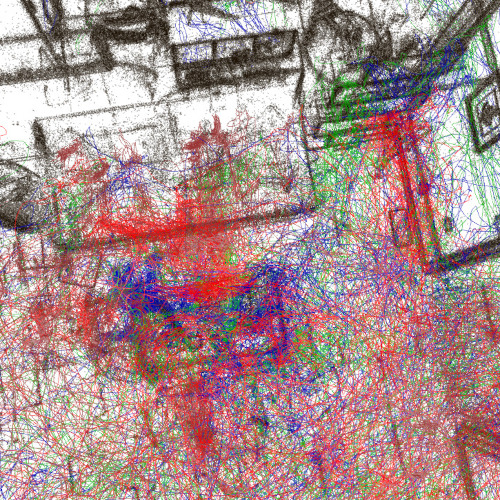}}%
        \end{minipage}~%
    }\\
    {
        \setlength{\fboxsep}{0pt}\setlength{\fboxrule}{0.5pt}
        \begin{minipage}{0.517\linewidth}%
        \fbox{\includegraphics[width=\linewidth]{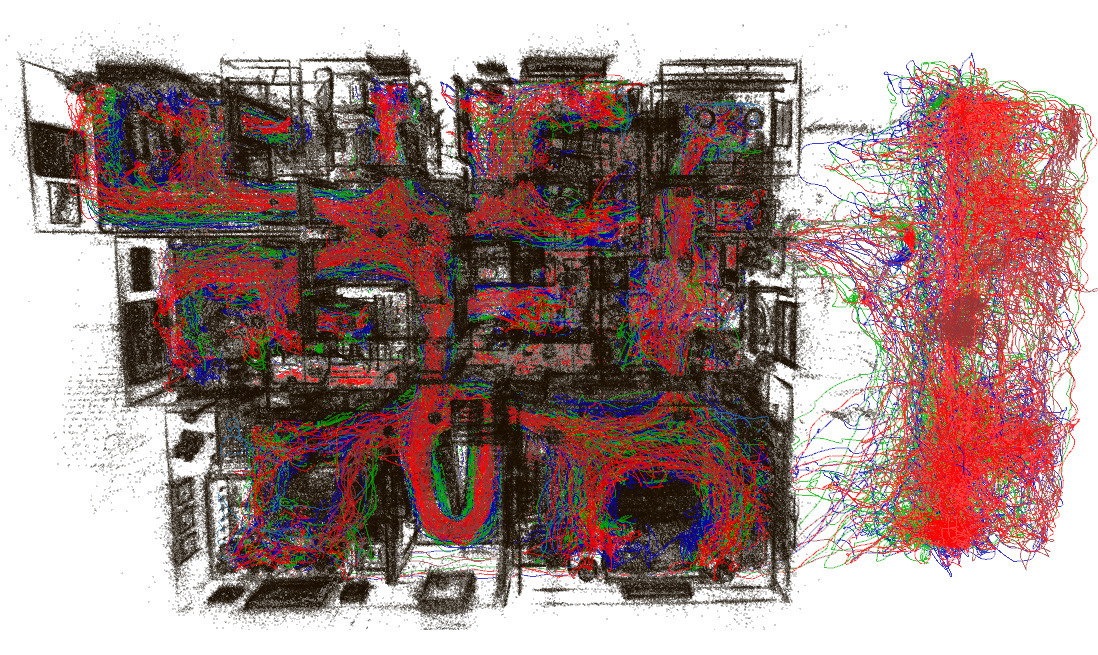}}%
        \end{minipage}~%
        \begin{minipage}{0.15\linewidth}%
        \fbox{\includegraphics[width=\linewidth]{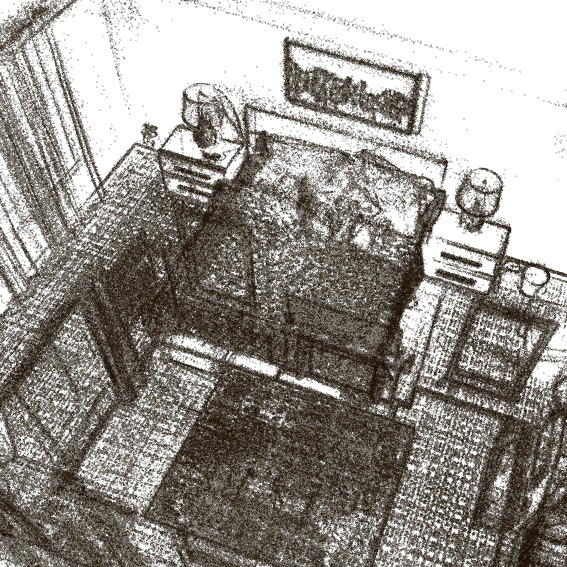}}\\[0.05mm]%
        \fbox{\includegraphics[width=\linewidth]{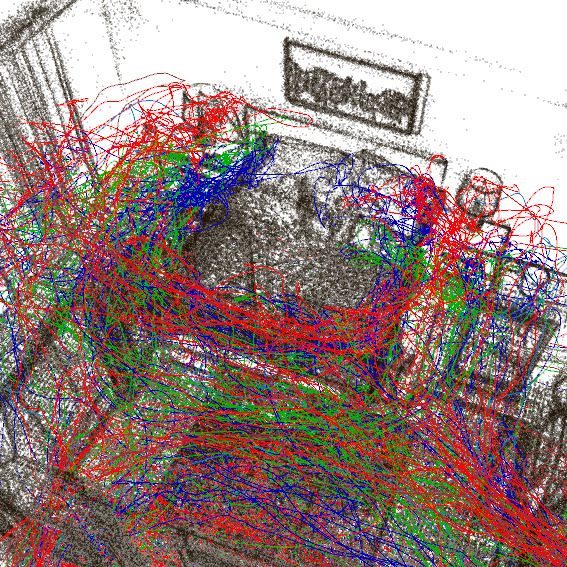}}%
        \end{minipage}~%
        \begin{minipage}{0.15\linewidth}%
        \fbox{\includegraphics[width=\linewidth]{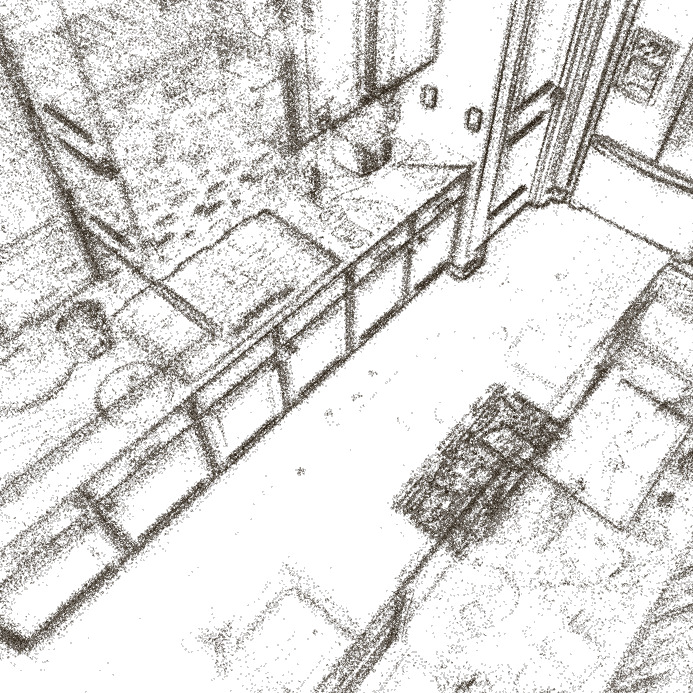}}\\[0.05mm]%
        \fbox{\includegraphics[width=\linewidth]{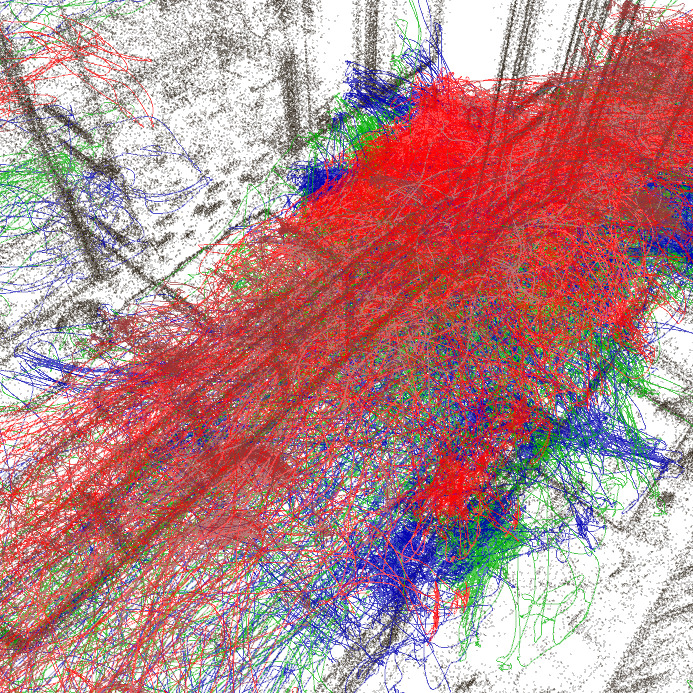}}%
        \end{minipage}~%
        \begin{minipage}{0.15\linewidth}%
        \fbox{\includegraphics[width=\linewidth]{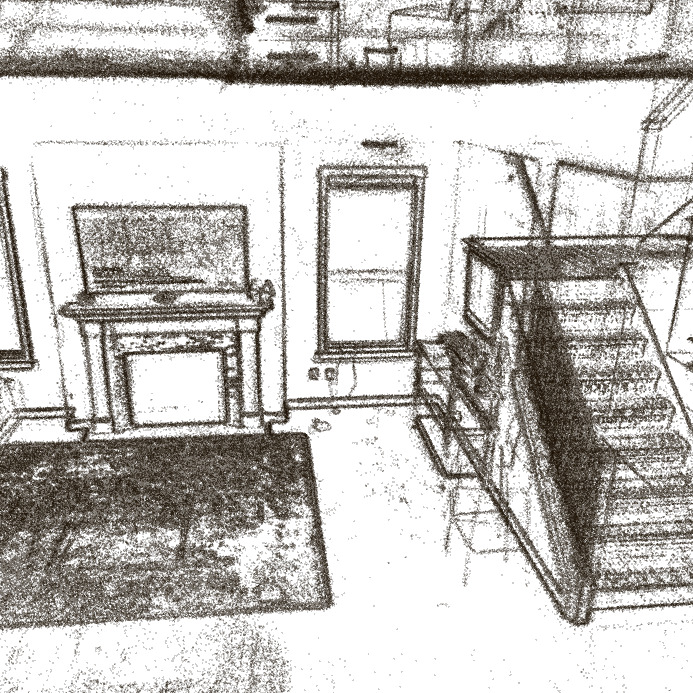}}\\[0.05mm]%
        \fbox{\includegraphics[width=\linewidth]{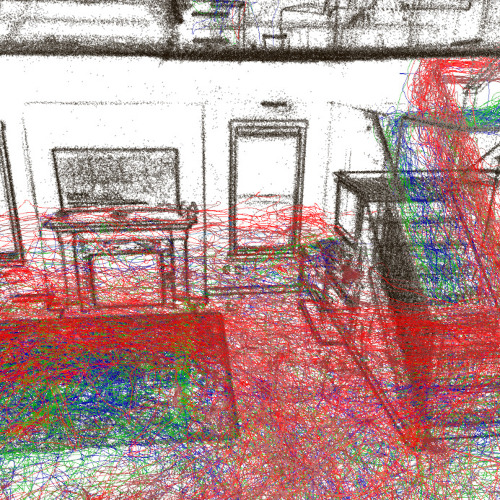}}%
        \end{minipage}~%
    }\\
    \caption{\textbf{Additional visualization of the MPS output for multiple recording locations from the \nymeria dataset}. Left: map and all trajectories aligned in it. Right: zoomed-in locations within the map, with and without trajectories. The observer and participant head trajectory is shown in red, and the left and right wrist trajectory are shown in green and blue respectively.}
    \label{fig:locationsA}
\end{figure}

The \nymeria dataset contains \loccnt locations, with \abcnt houses, 1 cafeteria with an outdoor patio, 1 multistory office building and 1 campus ground with parking lot and multiple hiking/biking trails. The last 3 locations covers different spaces of the same campus, where all recordings are aligned into the same world coordinates. To achieve this, we collected extra basemap recordings which connect the disjoint locations. We refer this merged location by BX. \Cref{tab:location-breakdown} present the details of each location, and the accumulated trajectory length per location. In addition to \cref{fig:airbnb} in the main submission, \cref{fig:locationsA} provides visualizations of more locations.

{
    \footnotesize
    \setlength{\tabcolsep}{2.pt}
    \newcolumntype{C}[1]{>{\centering\arraybackslash}p{#1}}
    \newcolumntype{R}[1]{>{\raggedleft\arraybackslash}p{#1}}
    \newcolumntype{L}[1]{>{\raggedright\arraybackslash}p{#1}}
    \begin{longtable}{L{11mm} R{6mm} R{9mm} *{3}{R{11mm}} R{6mm} R{11mm} *{2}{R{6mm} } R{16mm}}
    \toprule
loc.  &day  &dur./h   &H/Km     &L/Km     &R/Km    &m.f.         &type        &liv. &k\&d &other   \\  \endhead \toprule
AB02  &4    & 4.32    & 2.50    &  3.61   &  3.53  &\checkmark   &3b3b        &2    &1    &yd.la.ga.re. \\ \hline
AB03  &2    & 3.85    & 2.19    &  2.74   &  2.91  &\checkmark   &3b1b        &1    &1    &yd.la.   \\ \hline
AB04  &3    & 6.62    & 3.77    &  3.86   &  4.54  &             &2b2b        &1    &1    &yd.la.   \\ \hline
AB05  &2    & 3.87    & 2.51    &  3.02   &  3.22  &\checkmark   &2b1b        &1    &2    &yd.la.   \\ \hline
AB06  &3    & 4.94    & 4.35    &  4.84   &  5.58  &             &3b2b        &1    &1    &yd.\\ \hline
AB07  &2    & 4.38    & 3.35    &  5.74   &  5.87  &             &3b2b        &1    &1    &yd.of.   \\ \hline
AB08  &3    & 6.74    & 6.41    &  11.0   &  11.7  &\checkmark   &3b1b        &1    &1    &yd.la.of.  \\ \hline
AB10  &3    & 7.74    & 4.66    &  7.78   &  7.90  &             &1b1b        &1    &1    &yd.la.  \\ \hline
AB11  &4    & 10.8    & 8.20    &  12.8   &  13.2  &\checkmark   &3b2b        &1    &1    &yd.\\ \hline
AB12  &2    & 5.35    & 4.55    &  7.57   &  7.72  &\checkmark   &3b3b        &2    &2    &yd.la. \\ \hline
AB13  &3    & 8.06    & 8.61    &  14.9   &  15.2  &             &2b1b        &1    &2    & \\ \hline
AB14  &2    & 5.48    & 6.79    &  9.50   &  9.54  &             &3b2b        &1    &1    &yd.\\ \hline
AB15  &3    & 6.67    & 6.10    &  8.74   &  8.97  &             &3b2b        &1    &1    &yd.la.re.             \\ \hline
AB16  &2    & 5.46    & 5.98    &  9.24   &  9.41  &\checkmark   &5b3b        &2    &1    &yd.la.of.ga. \\ \hline
AB17  &3    & 7.47    & 6.69    &  11.7   &  11.0  &\checkmark   &4b2.5b      &1    &2    &\\ \hline
AB18  &2    & 5.26    & 4.96    &  7.67   &  7.67  &             &3b1b        &1    &1    &yd.\\ \hline
AB19  &2    & 5.20    & 3.97    &  6.50   &  6.70  &             &3b2b        &2    &2    &yd.la.ga.  \\ \hline
AB20  &3    & 6.80    & 5.92    &  8.93   &  9.58  &\checkmark   &4b3b        &1    &1    &yd.la.of.  \\ \hline
AB21  &2    & 5.45    & 4.69    &  8.36   &  8.31  &             &2b1b        &1    &2    &yd.la. \\ \hline
AB22  &3    & 7.88    & 8.15    &  11.8   &  12.5  &\checkmark   &4b2b        &1    &2    &yd.la. \\ \hline
AB23  &2    & 4.00    & 3.58    &  4.78   &  5.14  &             &3b2b        &1    &2    &yd.\\ \hline
AB24  &3    & 7.86    & 6.75    &  8.85   &  9.34  &\checkmark   &5b2b        &1    &2    &yd.\\ \hline
AB25  &2    & 5.35    & 5.72    &  6.79   &  7.16  &\checkmark   &4b2b        &2    &1    &yd.\\ \hline
AB26  &3    & 6.56    & 7.06    &  8.64   &  9.31  &\checkmark   &4b3.5b      &2    &4    &yd.la.ga.  \\ \hline
AB27  &2    & 3.95    & 3.93    &  4.60   &  5.07  &\checkmark   &9b2.5b      &1    &2    &yd.la.ga.re. \\ \hline
AB28  &2    & 5.27    & 5.21    &  5.99   &  6.63  &             &3b3b        &2    &2    &yd.la. \\ \hline
AB29  &3    & 8.01    & 8.68    &  10.9   &  11.2  &\checkmark   &4b2b        &1    &2    &yd.la. \\ \hline
AB30  &2    & 5.41    & 5.75    &  6.87   &  7.55  &\checkmark   &2b2b        &1    &2    & \\ \hline
AB31  &3    & 8.27    & 12.1    &  15.0   &  16.6  &\checkmark   &5b2b        &2    &2    &yd.la.ga.  \\ \hline
AB32  &2    & 5.12    & 7.09    &  7.76   &  8.91  &\checkmark   &4b3b        &1    &2    &yd.of.ga.         \\ \hline
AB33  &3    & 8.13    & 10.3    &  12.8   &  13.0  &\checkmark   &4b2b        &1    &2    &yd.re.             \\ \hline
AB34  &2    & 5.40    & 6.07    &  7.67   &  7.99  &\checkmark   &4b4b        &1    &2    &\\ \hline
AB35  &3    & 8.15    & 9.02    &  10.4   &  10.8  &\checkmark   &5b3b        &1    &2    &\\ \hline
AB36  &3    & 8.24    & 11.5    &  14.1   &  14.7  &\checkmark   &5b3b        &1    &2    &yd.la. \\ \hline
AB37  &2    & 5.45    & 6.91    &  8.24   &  9.08  &\checkmark   &4b2.5b      &1    &2    &yd.la.re.  \\ \hline
AB38  &3    & 8.28    & 10.1    &  11.8   &  13.4  &             &4b1.5b      &1    &2    &yd.re. \\ \hline
AB39  &2    & 4.11    & 3.75    &  4.28   &  5.11  &             &4b2b        &1    &2    &yd.la. \\ \hline
AB40* &4    & 15.3    & 14.0    &  16.7   &  17.8  &\checkmark   &4b3b        &2    &3    &yd.la.re.  \\ \hline
AB41  &3    & 7.74    & 8.08    &  9.40   &  10.9  &\checkmark   &4b3.5b      &1    &2    &yd.la.of.  \\ \hline
AB42  &2    & 5.45    & 5.82    &  6.60   &  8.03  &\checkmark   &4b3b        &1    &2    &yd.\\ \hline
AB43* &3    & 7.73    & 7.34    &  8.40   &  9.32  &             &5b2b        &2    &1    &yd.\\ \hline
AB44* &2    & 8.24    & 7.12    &  8.20   &  9.16  &\checkmark   &5b2b        &2    &3    &yd.\\ \hline
AB45* &3    & 12.9    & 9.20    &  11.6   &  12.7  &             &2b2b        &1    &2    &yd.la. \\ \hline
AB46* &2    & 8.52    & 6.76    &  8.01   &  8.73  &\checkmark   &6b3.5b      &1    &1    &yd.la.of.  \\ \hline
AB47* &3    & 11.8    & 12.1    &  15.8   &  17.0  &\checkmark   &4b2.5b      &1    &1    &yd.la.of.  \\ \hline
AB48* &2    & 6.30    & 4.95    &  5.96   &  6.78  &\checkmark   &3b2b        &1    &1    &yd. \\ \hline
AB49* &3    & 12.9    &12.98    & 15.88   &  16.8  &\checkmark   &3b3b        &1    &1    &yd.la.re.of. \\ \hline
BX*   &20   & 50.9    &94.53    &107.8    &110.9   &\checkmark   &            &     &     &\\ \midrule
sum   &122  &377.8    &411.0    &524.4    &554.4   &32           &            &58   &79   & \\
    \bottomrule
    \caption{\textbf{Summary of locations.} For each location, we report the number of days spent onsite, the recorded data duration in hour, the trajectory lengths in Km of head(H), left wrist(L) and right wrist(R), respectively. The table also summarizes the location layout. The prefix AB indicates houses and the location BX combines 3 areas on a large campus. Locations marked with * contain two-participant collections. The abbreviations have the following meanings: XbYb for X bedrooms and Y bathrooms, m.f. for multi-floor, liv. for living room, k\&d for kitchen and dinning space, yd. for yard, la. for laundry room, ga. for garage, re. for recreation room, and of. for office room. Note the summed hour is longer than \hourcnt for the same reason given in Table~\ref{tab:scenario}.}
    \label{tab:location-breakdown}
    \end{longtable}
}

\section{Benchmarks}\label{supp:benchmark}
In this section we provide details about how baseline methods are trained and evaluated with our dataset. We also provide qualitative results of the baseline algorithms in addition to the quantitative evaluations reported in the main submission.

\subsubsection{AvatarPoser.}
AvatarPoser~\cite{avatarposer22eccv} is a regressive method based on the transformer architecture. It takes positional information about 3 body joints (head and hands) for the last 40 frames (0.66\,s) and outputs full-body pose prediction for the last frame in a form of local rotations for each joint. The translation of the body is then inferred by using the input head joint position and traversing it up the kinematic chain to the model's root joint. The model does not predict the shape of the body and uses GT shape parameters to infer the position of the joints.
The input positional information is encoded as a vector of the translation, rotation and linear and angular velocity of head and hands.
Original model assumes SMPL~\cite{loper15smpl} body model; to benchmark the method on our dataset, we adapted the inference pipeline to work with the Xsens kinematic tree and increased the number of predicted joints from 22 to 23; otherwise, we tried to maintain the original pipeline intact, reusing the original preprocessing and training scripts, and retrained the model from scratch.

\subsubsection{BoDifffusion.}
BoDiffusion~\cite{bodiffusion2023iccv} is a 3-point to full-body motion generation method based on a Diffusion framework. The method takes a sequence of 3 point information frames and produces a window of full-body motions. BoDiffusion uses same input and output representations as AvatarPoser, sharing similar data preprocessing and motion inference pipelines. The architecture used in BoDiffusion is DiT~\cite{peebles2023scalable} -- a Transformer-based architecture with global conditioning. Similar to AvatarPoser, we only modified the inference method to work with the Xsens skeleton and changed the output size, keeping other parts original. We retrained the model from scratch on our data -- it took about 3 times longer to converge on our data compared to AMASS.

\subsubsection{EgoEgo.}
EgoEgo~\cite{egoego23cvpr} method is a diffusion-based method which predicts the full-body motion from head-mounted camera images. It has a two-stage pipeline: the first stage is a visual localization network, predicting the head trajectory from a series of camera images. The second part is a diffusion-based method which generates the full-body motion given the head trajectory predicted on the previous stage. Since the localization of our SLAM system is very robust, we are only interested in benchmarking the performance of the full-body prediction stage, therefore we dropped the visual localization stage and supplied the head positions to the diffusion-based model directly. Similar to the other baselines, we adapted the method to work with Xsens skeleton instead of SMPL, keeping the original architecture of the model. To train the model from scratch we used BoDiffusion diffusion training pipeline and code base as it was easier to adapt for our data.

\subsubsection{VQ-VAE.}
VQ-VAE~\cite{vq2017neurips, dhariwal2020jukebox} encodes high-dimensional data into discrete latent codes. With the large amount of motion data in this dataset, we are able to train a VQ-VAE that encodes motion representations into sequences of motion tokens, which has wide applications in motion generation and understanding~\cite{guo2022tm2t,motiongpt24neurips,motiongpt23arxiv}. To train the VQ-VAE, we first represent human motions as sequences of poses and root joint translation velocity and rotation. Each pose is represented by joint angles and joint rotation velocity defined on a kinematic tree. The motion VQ-VAE consists of fully convolutional encoder and decoder, which makes it capable of processing motions with arbitrary lengths. Both the encoder and decoder comprise three layers of 1-D convolution residual blocks. For the latent quantization, we use three techniques, including exponential moving average, codebook reset~\cite{dhariwal2020jukebox} and product quantization~\cite{lucas2022posegpt}, to improve the codebook usage rate and expressiveness, which are important to the reconstruction performance. For training, we use motion batches with window size of 1 second, which is equivalent to 60 frames under 60 fps.

\subsubsection{TM2T.}
TM2T~\cite{guo2022tm2t} is one of the early attempts that encodes motions into discrete tokens and use transformers for the generation and understanding of motions. The method first trains a motion VQ-VAE to map contiguous motion representations into discrete motion tokens. Then a transformer-based architecture is trained to map motion tokens to language tokens for motion understanding, or the other around for text-to-motion generation. To benchmark the method on our dataset, we use the motion VQ-VAE described above as the motion tokenizer. We adopt GPT-2~\cite{gpt2} tokenizer for motion narration tokenization. For each motion segment, we stack the motion narration for full body, upper body and lower body together to form the target motion narration. We are interested in the motion understanding part of the work. Therefore, we train the motion-to-text generation as the motion understanding baseline, using the offcial codes.

\subsubsection{MotionGPT.}
MotionGPT~\cite{motiongpt24neurips} uses language models to establish the joint distribution of motion and language so that the model can be prompted with natural languages for different motion tasks, \eg, motion prediction, motion in-between, text-to-motion generation, motion understanding. Similarly, MotionGPT also trains a motion VQ-VAE to tokenize the motions. Then to establish the joint distribution, MotionGPT starts from a pre-trained language model T5~\cite{raffel2020exploring} and trains with motion-to-text and text-to-motion translation tasks as the second stage. Lastly, instruction tuning is performed to prompt the model for different downstream tasks. Similar to TM2T, we test motion understanding on MotionGPT with our dataset. We use the same VQ-VAE described above as the motion tokenizer. We use the official codes and perform the second stage training with motion-to-text and text-to-motion tasks. Then we directly use the model from this stage to test motion understanding.

\begin{figure}[tb]
  \centering
  \includegraphics[width=\textwidth]{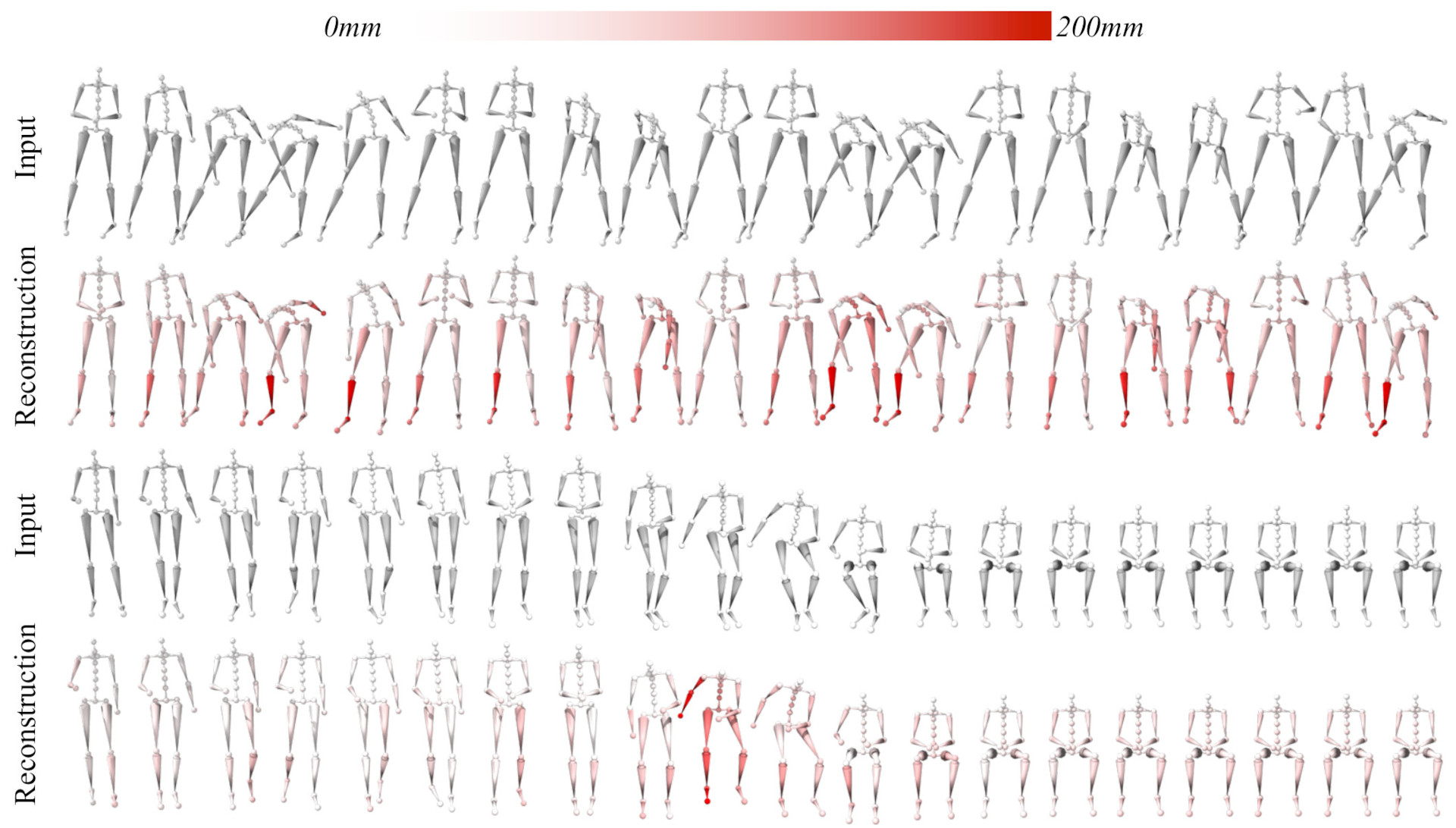}
  \caption{\textbf{VQ-VAE motion reconstruction.} This visualization is produced by the best VQ-VAE we trained, which corresponds to the last line of main paper Tab.~4.}
  \label{fig:supp_motion_vqvae}
\end{figure}

\begin{figure}[tb]
  \centering
  \includegraphics[width=\textwidth]{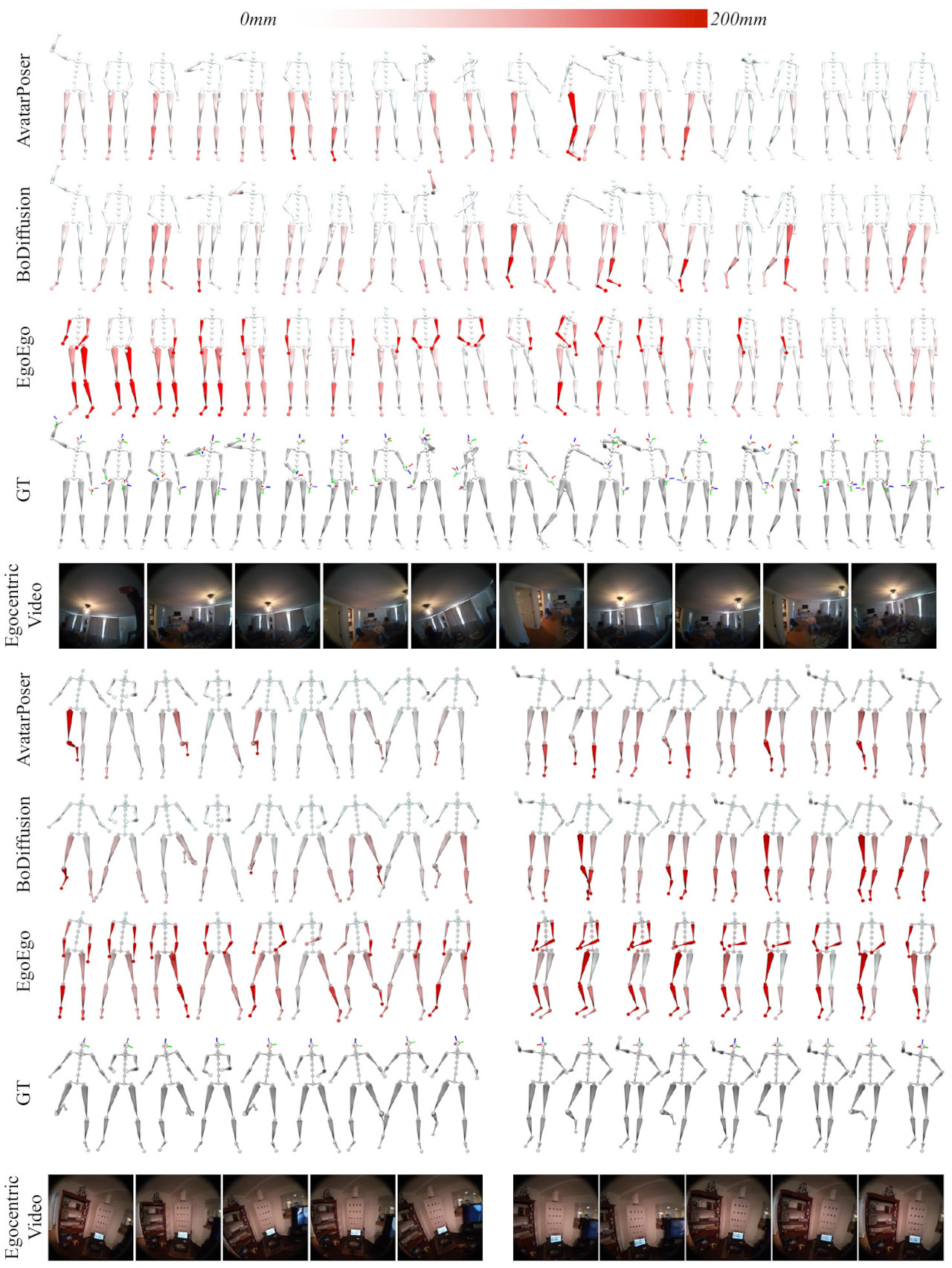}
  \caption{\textbf{Full-body tracking from 3-point and 1-point motion input.}}
  \label{fig:supp_motion_tracking}
\end{figure}

\begin{figure}[tb]
  \centering
  \includegraphics[width=\textwidth]{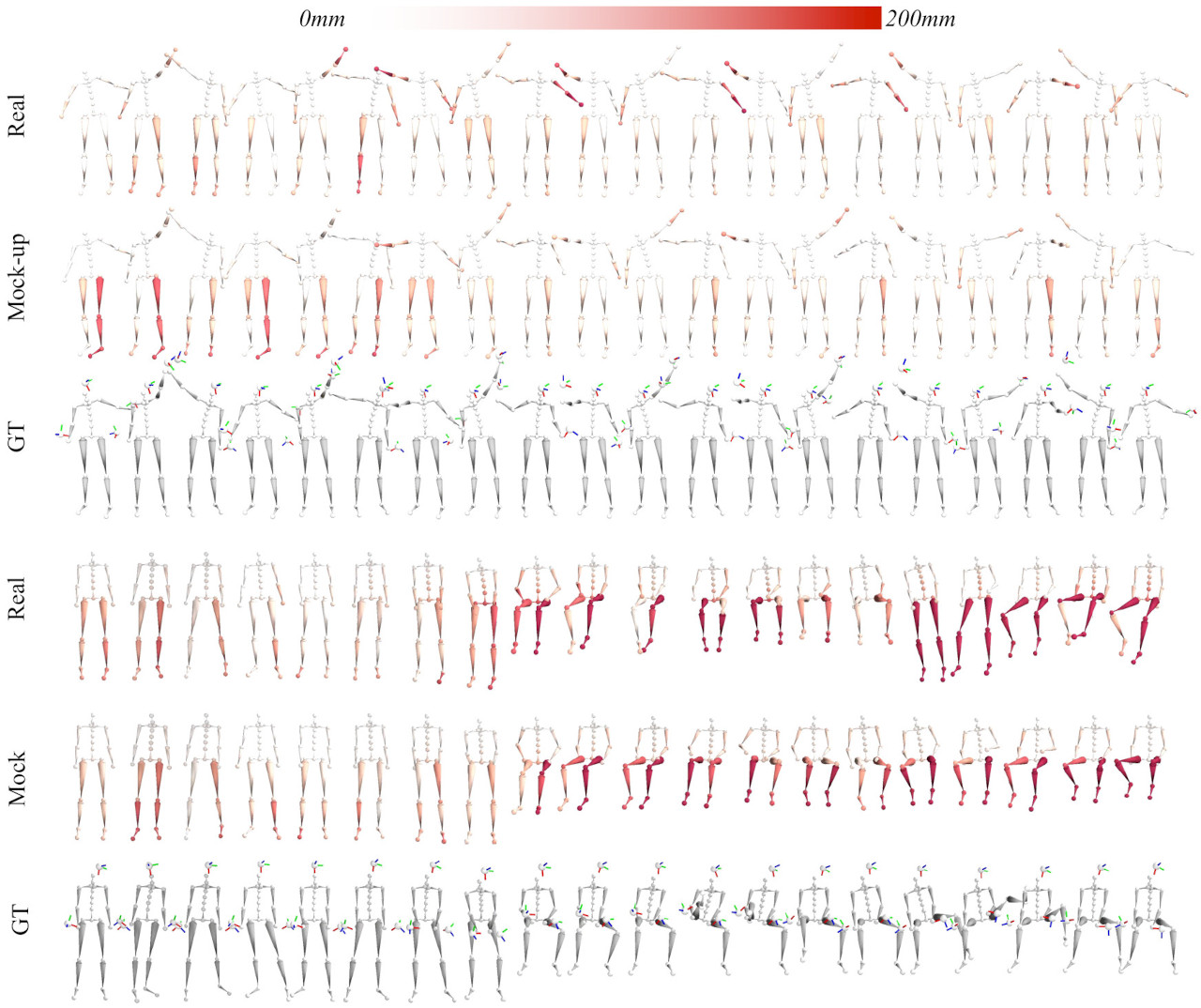}
  \caption{\textbf{Full-body tracking from real 3-points v.s. mock-up 3-points.}}
  \label{fig:supp_motion_tracking_real_vs_mockup}
\end{figure}

\begin{figure}[tb]
  \centering
  \includegraphics[width=\textwidth]{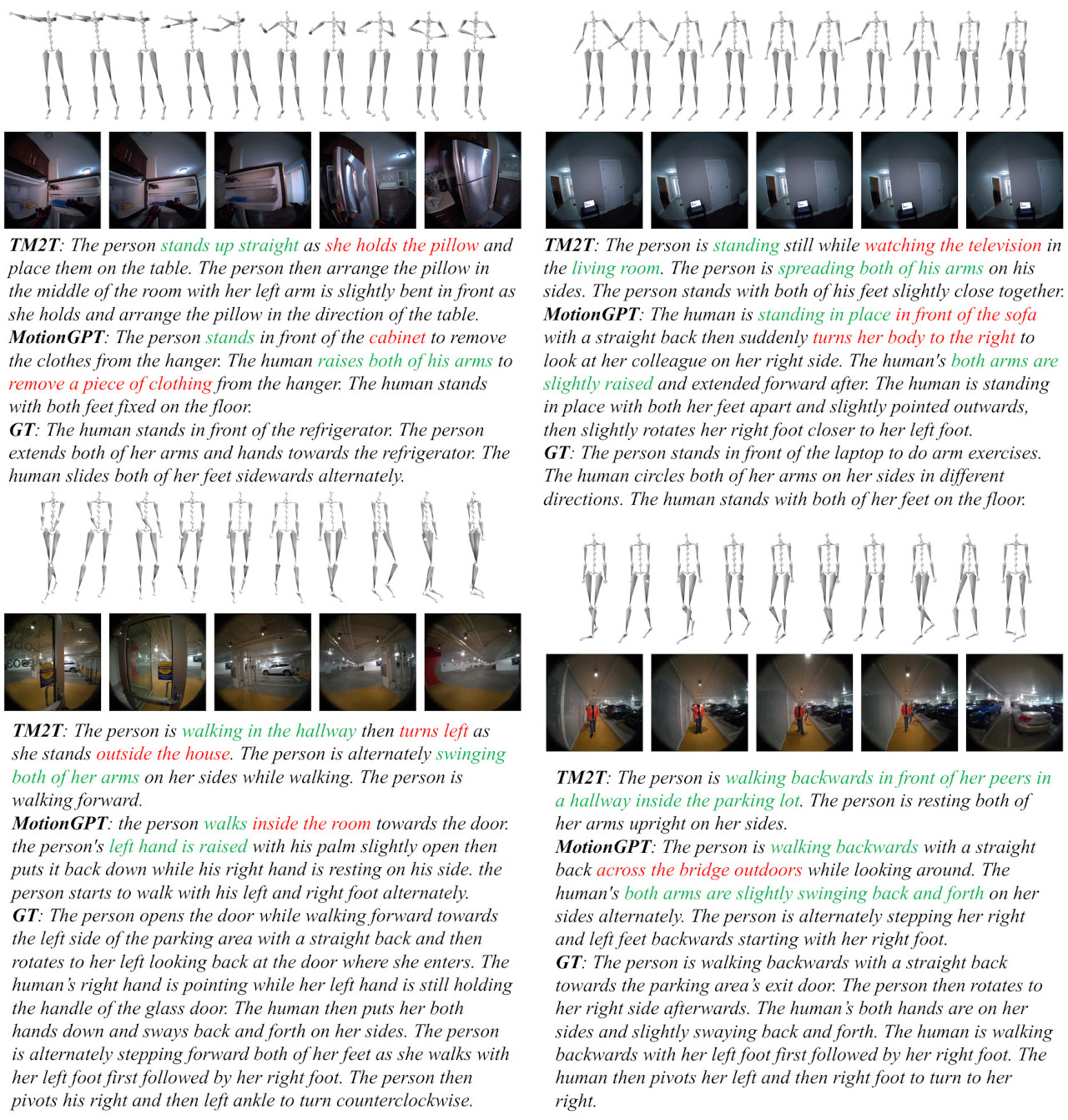}
  \caption{\textbf{Motion understanding from 3-point input.} }
  \label{fig:supp_motion_understanding}
\end{figure}

% ---- Bibliography ----
\clearpage
\bibliographystyle{splncs04}
\bibliography{main}

\end{document}